\documentclass[letterpaper, 10 pt, conference]{ieeeconf}  

\IEEEoverridecommandlockouts                              

\usepackage{graphics} 
\usepackage{epsfig} 
\usepackage{mathptmx} 
\usepackage{times} 
\usepackage{amsmath} 
\usepackage{amssymb}  
\usepackage{pifont}
\usepackage{subcaption}
\usepackage[export]{adjustbox}
\usepackage{multicol}
\usepackage{multirow}
\usepackage{tabularx}
\newcolumntype{Y}{>{\centering\arraybackslash}X}%
\usepackage{hhline}
\usepackage{makecell}
\usepackage{colortbl}
\usepackage[dvipsnames]{xcolor}
\usepackage{threeparttable}

\newcommand{\etal}[0]{\textit{et~al.}}

\title{\LARGE \bf
Markerless Robot Detection and 6D Pose Estimation for \\
Multi-Agent SLAM 
}

\author{Markus R\"uggeberg$^{1}$, Maximilian Ulmer$^{1, 2}$, Maximilian Durner$^{1}$,\\
Wout Boerdijk$^{1}$, %
Marcus M\"uller$^{1}$, Rudolph Triebel$^{1,2}$ and Riccardo Giubilato$^{1\dag}$
\thanks{\dag \ Corresponding author}
\thanks{*This work was supported by the Helmholtz Association project iFOODis (contract number KA2-HSC-06) and by the German Federal Ministry of Research, Technology and Space (BMFTR) under the Robotics Institute Germany (RIG)}
\thanks{$^{1}$Institute of Robotics and Mechatronics, German Aerospace Center (DLR), Weßling, Germany
        {\tt\small fistname.lastname@dlr.de}}%
\thanks{$^{2}$Rudolph Triebel is also with Karlsruhe Institute of Technology (KIT), Karlsruhe, Germany
        {\tt\small rudolph.triebel@kit.de}}%
}

\usepackage{xcolor}
\usepackage{makecell}

\definecolor{dgreen}{rgb}{0.2, 0.6, 0}
\newcommand{\dgreen}[1]{\textcolor{dgreen}{#1}}
\definecolor{dred}{rgb}{0.75, 0, 0}
\newcommand{\dred}[1]{\textcolor{dred}{#1}}
\definecolor{LRU_blue}{HTML}{007FDF}
\definecolor{Lander_green}{HTML}{00BF00}

\begin{document}

\maketitle
\thispagestyle{empty}
\pagestyle{empty}

\begin{abstract}
The capability of multi-robot SLAM approaches to merge localization history and maps from different observers is often challenged by the difficulty in establishing data association. Loop closure detection between perceptual inputs of different robotic agents is easily compromised in the context of perceptual aliasing, or when perspectives differ significantly. For this reason, direct mutual observation among robots is a powerful way to connect partial SLAM graphs, but often relies on the presence of calibrated arrays of fiducial markers (e.g., AprilTag arrays), which severely limits the range of observations and frequently fails under sharp lighting conditions, e.g., reflections or overexposure. In this work, we propose a novel solution to this problem leveraging recent advances in Deep-Learning-based 6D pose estimation. We feature markerless pose estimation as part of a decentralized multi-robot SLAM system and demonstrate the benefit to the relative localization accuracy among the robotic team. The solution is validated experimentally on data recorded in a test field campaign on a planetary analogous environment.  
\end{abstract}

\section{INTRODUCTION}
A necessary capability of robotic teams to perform collaborative actions, e.g., industrial inspection \cite{tavakoli2012cooperative}, rescue \cite{queralta2020collaborative}, or scientific exploration \cite{schuster2020arches}, is the one of mutual localization and collaborative mapping. Collaborative, or multi-robot Simultaneous Localization and Mapping approaches (SLAM), extend the ability of single robots to create maps of unknown environments, and localize themselves with respect to it, to multiple agents operating at the same time. 

This conceptually simple addition brings a variety of complications to the table, from the perspective of many components of a traditional SLAM system. 
From an architectural point of view, multiple strategies to fuse data from the robots have been explored. Centralized SLAM approaches \cite{schmuck2019ccm} vs. distributed SLAM approaches \cite{lajoie2020door} bring considerable consequences both in term of algorithmic complexity as well as system behaviour and robustness to failure during operation. From the perceptual point of view, the task of associating measurements between robots, i.e., performing loop closures in visual SLAM systems, is severely affected by strong perspective and illumination changes, due to viewpoint differences between the systems \cite{giubilato2022challenges}. Similarly, the problem of data association in LiDAR-based perception systems is affected by different noise patterns and pointcloud sparsity when robots operate at a distance. These issues are further exacerbated when operating on natural unstructured grounds, e.g., in the context of field robotic operations or planetary exploration, where no particular constraints are imposed on traversable paths. 

\begin{figure}[!tp]
    \centering
    \begin{subfigure}{\linewidth}
    \includegraphics[trim={0 8cm 0 0}, clip, width=\linewidth]{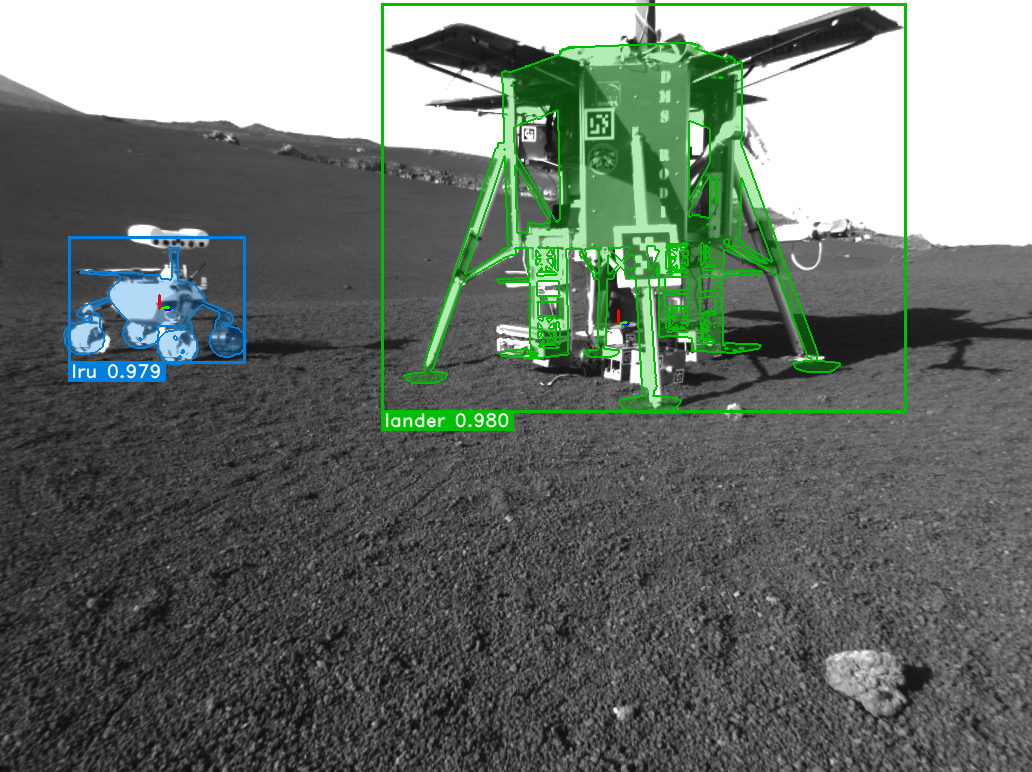}
    \end{subfigure} \\
    \vspace{.3em}
    \begin{subfigure}{\linewidth}
    \includegraphics[width=\linewidth]{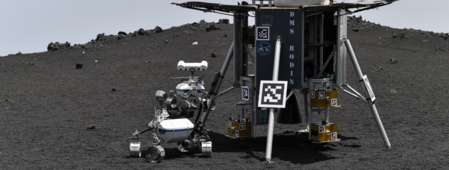}
    \end{subfigure}
    \caption{(\textbf{top:}) Multi-robot detection of the {\color{LRU_blue}LRU} (Lightweight Rover Unit), and the {\color{Lander_green}Lander} unit, from the perspective of the 2nd LRU unit during the 
    [anonymous]
    field test campaign on Mount. Etna, Sicily. The figure shows the projection of the robot shapes, known a-priori, to demonstrate the quality of pose estimation. The distance of the {\color{LRU_blue}LRU} rover to the observer, the LRU2, as well as intense light reflections, would make it impossible through conventional fiducial markers to establish a robot detection. (\textbf{bottom:}) Members of the multi-robot team on Mt. Etna: LRU, LRU2 and the Lander unit.}
    \label{fig:arches_example_pose_estimation}
\end{figure}

\begin{table*}[!tp]
\caption{Comparison of Selected Collaborative SLAM, Tracking or Localization Approaches by Inter-Robot Constraint}
\label{tab:related_work}
\centering
\renewcommand{\arraystretch}{1.2}
\newcommand{\cmark}{\ding{51}}%
\newcommand{\xmark}{\ding{55}}%
\begin{threeparttable}
\begin{tabularx}{\linewidth}{lXcccc}
\hline
\textbf{Method} & \textbf{Constraint Source} & \textbf{Detects Robots} & \textbf{Use Markers} & \textbf{Detection Type} & \textbf{Modality} \\
\hline 
Kimera-Multi \cite{rosinol2021kimera} & VPR, BoW vectors & \xmark & - & - & Visual \\
DOOR-SLAM \cite{lajoie2020door} & VPR, NetVLAD descriptors & \xmark & - & - & Visual \\ 
CVI-SLAM \cite{karrer2018cvi} & VPR, visual features & \xmark & - & - & Visual \\
CCM-SLAM \cite{schmuck2019ccm} & VPR, visual features & \xmark & - & - & Visual \\
LAMP 2.0 \cite{chang2022lamp} & LiDAR scan registration & \xmark & - & - & LiDAR \\
DiSCo-SLAM \cite{huang2021disco} & Scan Context descriptors & \xmark & - & - & LiDAR \\
Swarm-SLAM \cite{lajoie2023swarm} & Scan Context descriptors & \xmark & - & - & LiDAR \\ \hline 
MultiReg \cite{5354560} & Generic measurements & \cmark & \cmark \tnote{\dag} & SE(2) & Generic \\
UVDAR \cite{petravcek2020bio} & Direct robot measurements & \cmark & \cmark (UV LEDs) & Range/Bearing & Visual \\
CREPES \cite{xun2023crepes} & Camera, IMU and UWB fusion & \cmark (LEDs) & \cmark & SE(3) & Multiple \\
\cite{steidle2023temporal} & Camera tracking \& target VIO fusion & \cmark & \xmark & SE(3) & Visual \\
\cite{jin2019drone} & DL-based keypoint extraction & \cmark & \xmark & SE(3) & Visual \\
MSL-Raptor \cite{ramtoula2020msl} & DL-based bounding-box extraction & \cmark & \xmark & SE(3) & Visual \\
\hline
\end{tabularx}
\begin{tablenotes}
\item[\dag] Markers are specific geometries, e.g., square plates, but missing unique identifiers 
\end{tablenotes}
\end{threeparttable}
\end{table*}

Association between measurements from the robots in the team is of critical importance in collaborative SLAM, as it allows robots to join their environment representations into a single, and shared, formulation. Under aliased and ambiguous perceptual conditions, correspondence between views of the environments from multiple agents is challenging, therefore \textbf{direct observation} of other robots allow to establish immediate correspondences between, e.g., SLAM graphs, solving the association problem in a relatively straight-forward manner. While many existing collaborative SLAM approaches do not explicitly consider this case, the ones that do often utilize traditional visual markers, e.g. AprilTag \cite{wang2016apriltag} or ArUco \cite{garrido2014automatic}. This choice, however, brings significant limitations: the size of the markers limits the \textbf{maximum distance} at which a robot can be detected, the mounting positions limits \textbf{relative orientations} valid for detection, and \textbf{harsh lighting} conditions in outdoor contexts can prevent detection entirely.

We propose a solution to this problem taking inspiration from recent advances in 6D pose estimation techniques for arbitrary objects \cite{ulmer20236d}. With the assumption that the robots that constitute a team have a known shape and appearance, we employ a deep learning-based approach to detection and relative 6D pose estimation. A simplified prior shape of each robot is aligned to detected instances of the robots in each other's camera image, estimating a full transformation that is directly used in a SLAM context. In summary, this manuscript presents the following contributions: 
\begin{itemize}
    \item To the best of our knowledge, we present for the first time a \textbf{markerless detection and pose estimation} approach to mutual robot detection, that relies on prior shape knowledge without handcrafted features or visual markers.
    \item We demonstrate the integration of our markerless robot detection scheme in an online \textbf{multi-robot SLAM system} for a team of planetary rovers.
    \item We highlight the performance of our system, and the benefit against traditional marker-based pose estimation on a large body of data recorded during a test campaign on a planetary analogous environment.
    \end{itemize}

\section{Related Works}
In this section we provide an overview of related multi-agent SLAM works, regardless of the principal sensing modality, with the focus on methods that put the emphasis on the perceptual challenge in associating measurements from the members of the robotic team. Among these, there exist methods that consider direct mutual measurements between the agents, such as our proposed approach, or try to associate measurements from the environment shared across the team. Table~\ref{tab:related_work} presents a synthetic overview of the works that are discussed in the next sections. 

\begin{figure*}[!tp]
\centering
\includegraphics[width=\textwidth]{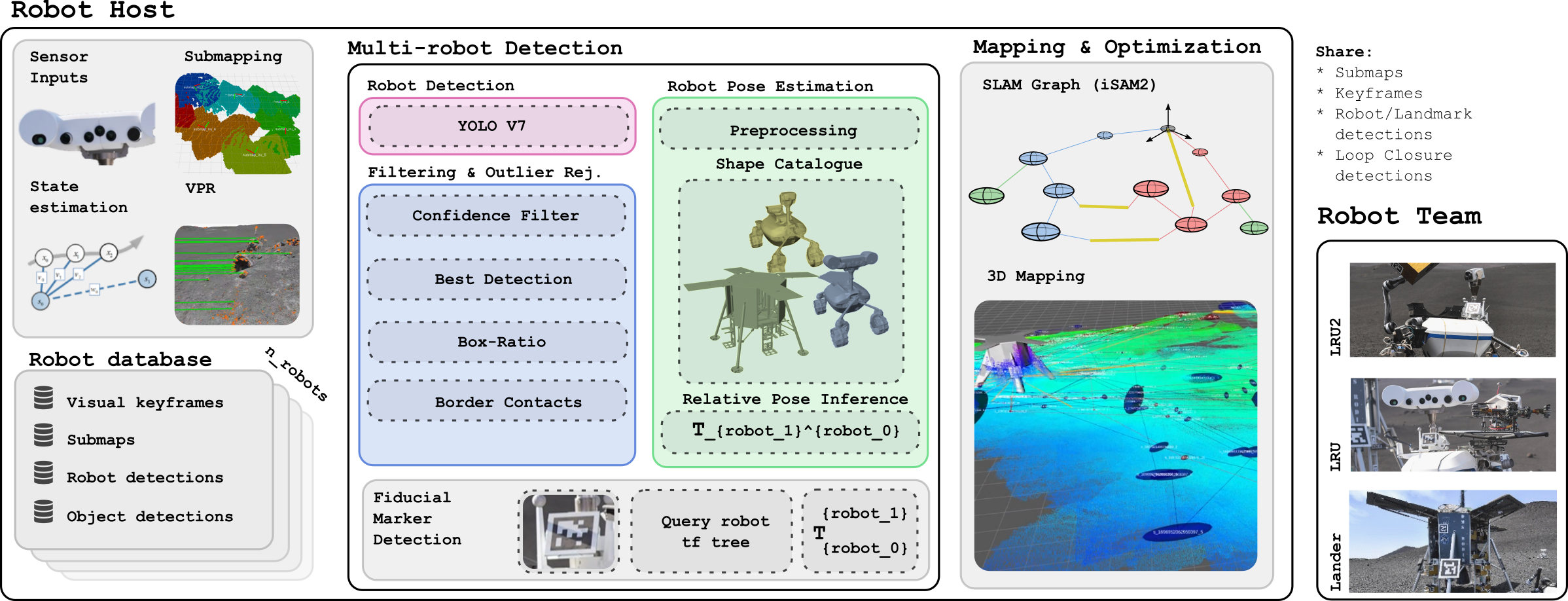}
\caption{Schematic overview of the employed decentralized SLAM system, with a focus on the multi-robot detection module capabilities. Each robot utilizes visual and inertial inputs to compute state estimation and partition robot states into submaps. Visual inputs are used to compute visual keyframes for place recognition, as well as detecting robots in the image, either through legacy AprilTag detection, or the proposed markerless approach. Results of submapping and robot detections from each robot in the team are embedded into a SLAM graph} 
\label{fig:architecture}
\end{figure*}
\subsection{Indirect Multi-Robot Associations}
In this category belong multi-agent SLAM systems that do not explicitly consider direct robot detection to join estimates from the team participants, but instead focus on the detection of \textit{loop closures}, i.e., the association of measurable perceptual properties of the environment. Kimera-Multi \cite{tian2022kimera} is a distributed multi-robot SLAM system that extend the visual-inertial and semantic mapping framework Kimera \cite{rosinol2021kimera}. The multi-robot team shares Bag-of-Words vectors computed from traditional image features and attempts to find consistent associations using an incremental formulation of PCM (Pairwise Consistent Measurements), followed by a robust optimization scheme. Similar to Kimera-Multi is DOOR-SLAM \cite{lajoie2020door}, a decentralized multi-robot approach that too relies on careful selection of loop closure candidates through PCM. Differently from Kimera-Multi, however, visual keyframes are processed to extract global descriptors with NetVLAD, validating putative matches with geometrical verification. The collaborative visual SLAM CVI-SLAM \cite{karrer2018cvi}, and its evolution CCM-SLAM \cite{schmuck2019ccm} are centralized approaches where a number of agents run only a visual-inertial \textit{front-end} and communicate keyframes and feature descriptors to a central server, which runs a loop closure detection and mapping engine, localizing agents on a shared feature map. All these methods rely on a visual modality to find correspondences between environment observations, which is prone to failure when the appearance is aliased and ambiguous \cite{giubilato2022challenges}. Another family of multi-robot SLAM approaches focus instead on the LiDAR modality, emphasizing structure rather than texture. LAMP2.0 \cite{chang2022lamp} is a centralized LiDAR-based multi-agent SLAM approach where a central server maintains and joins pose graphs from the participating agents based on scan registration. As the authors mention, however, the correspondences are prioritized based on an initial pose prior, which must be obtained by heuristics. LiDAR-based loop closure detections are instead used by the distributed SLAM approach DiSCo-SLAM \cite{huang2021disco}. By means of sharing and matching Scan Context descriptors \cite{kim2018scan}, robots attempt to initialize and register their maps among a team. Although LiDAR sensing benefits from the robustness and invariance of structure to different perceptual conditions, the absence of salient structure, such as in the case of planetary-like scenes, rapidly leads to failures, as the authors of Swarm-SLAM \cite{lajoie2023swarm} demonstrate in their field tests \cite{lajoie2025multi}.

\subsection{Direct Multi-Robot Associations}
Directly measuring transformations between robots using onboard perception has been investigated for several purposes, either tracking systems from a base station, to provide absolute localization with respect to it, or to enable relative localization capabilities to a team of robots in the context of SLAM. 
The authors of \cite{5354560} investigate the topic of relative localization between robots by means of mutual detection. Their system, named MultiReg, tackles the case where robot identities in a swarm are either ambiguous or unknown, and develop a theoretical framework to establish and validate detections. The approach is demonstrated with LiDAR-equipped systems, and, while theoretically applicable to different perception systems, visual fiducial markers could easily provide means of identification.
Other approaches rely on visual perception systems and fiducial markers, which provide both an identity and means to establish transformations. The authors of \cite{petravcek2020bio} present UVDAR, a biologically inspired method for swarm localization among a team of UAVs. The author solve the problem of mutual localization between systems by means of observing LED patterns with specific blinking frequencies with UV-sensitive cameras. Similarly, the authors of CREPES \cite{xun2023crepes} propose a relative localization system that rely on LEDs to define an identity, while using mutual state estimation inputs to compute relative poses. In order to avoid implementing specific hardware or preparing robots with fiducial markers, which especially for small agile system can be undesirable, the authors of MSL-Raptor \cite{ramtoula2020msl} propose a Deep Learning architecture to infer the coordinates of a bounding box around a target drone observed from a camera. Similarly, the authors of \cite{jin2019drone} propose to learn instead drone-specific visual keypoints pointing to features such as rotor position. 
These approaches show multiple alternatives to detect and estimate the pose of known robots, but either require specific hardware, e.g. LEDs, or consider specific robot features that do not easily generalize. With this work we aim to provide a general solution to this problem that extends to arbitrary robots equipped with vision sensors, and for arbitrary contexts of operation. 

\section{Methods}
In this section, we present the methods and components of our work, consisting of a multi-robot SLAM system (\ref{sec:multirobot_slam}), a learning-based markerless robot detection and 6D pose estimation pipeline (\ref{sec:pose_est}), and a systematic training framework with planetary analogue synthetic training data (\ref{sec:training_process}).

\begin{figure*}[!t]
  \centering
  \begin{subfigure}{0.24\textwidth}
    \includegraphics[width=\linewidth]{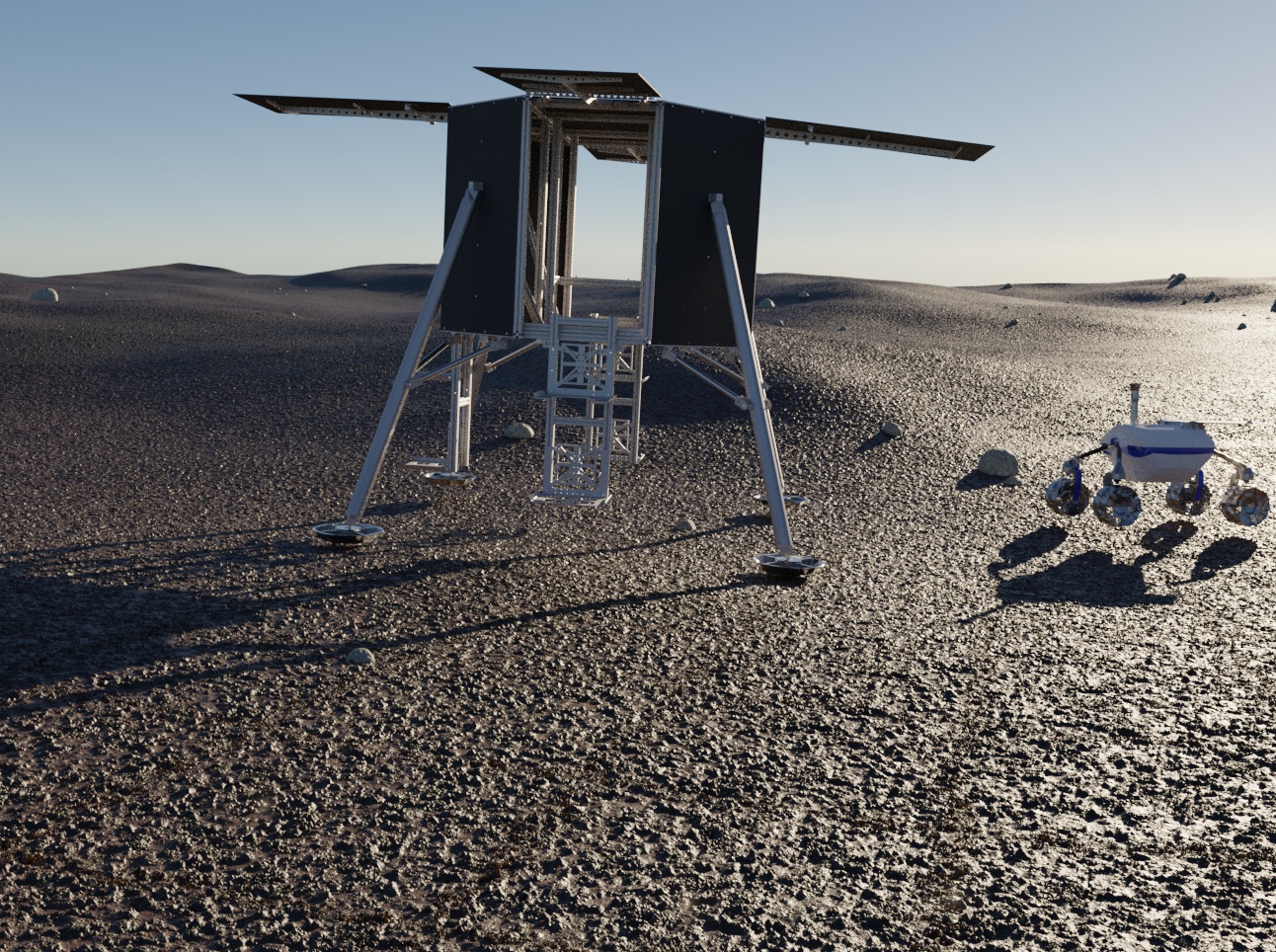}
  \end{subfigure}\hspace{0.1em}
  \begin{subfigure}{0.24\textwidth}
    \includegraphics[width=\linewidth]{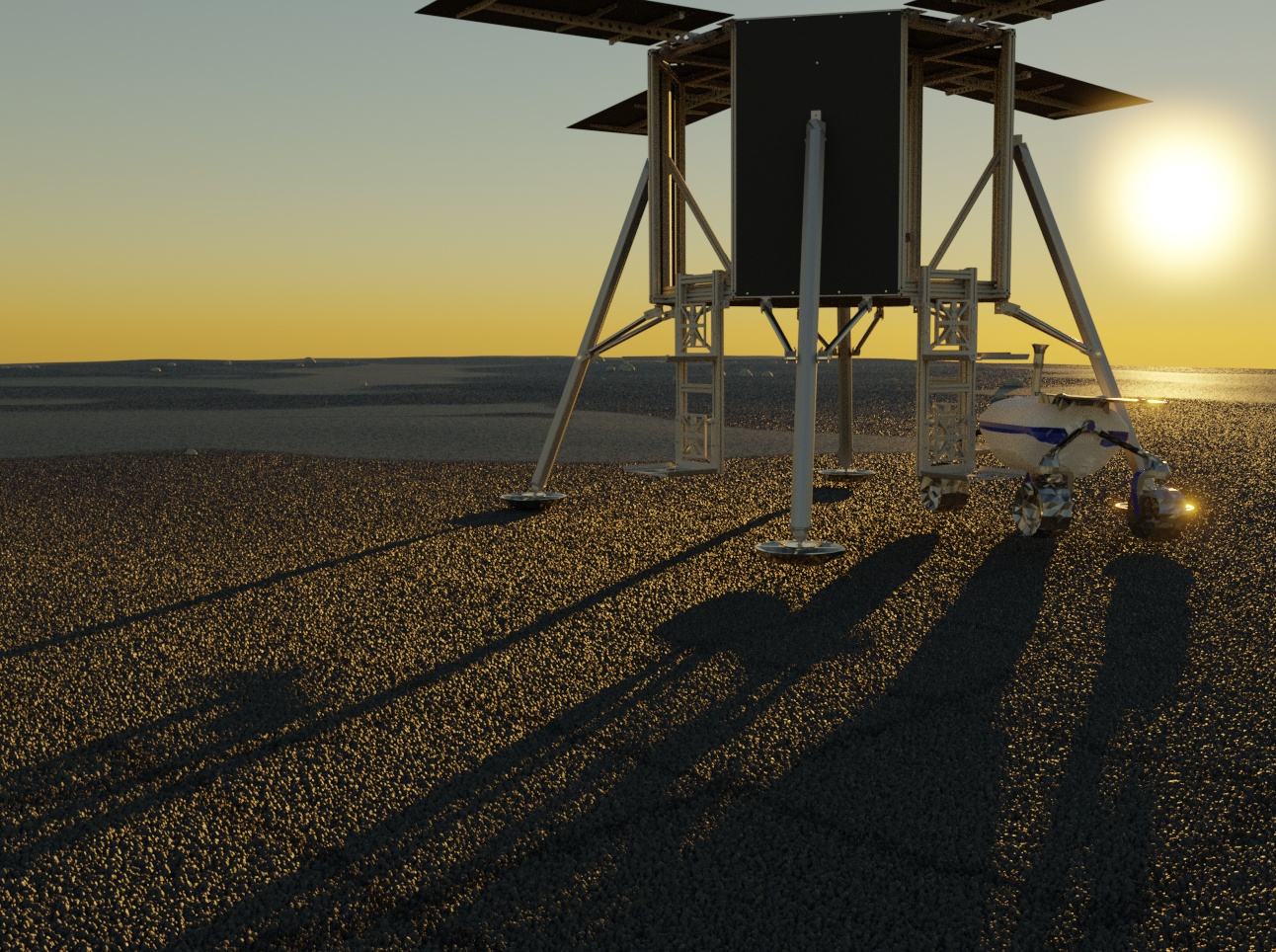}
  \end{subfigure}\hfill
   \begin{subfigure}{0.24\textwidth}
   \includegraphics[width=\linewidth]{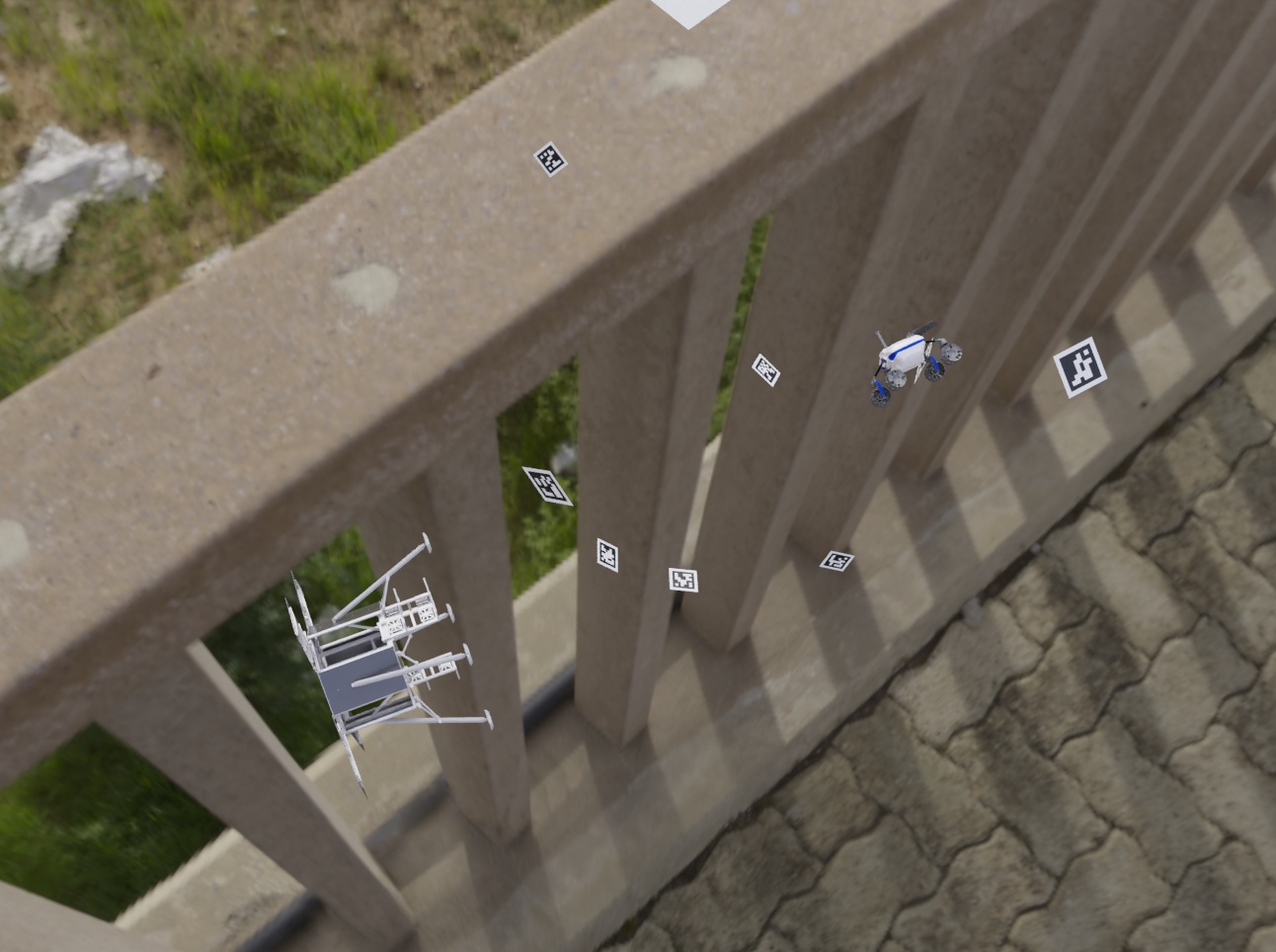}
 \end{subfigure}\hspace{0.1em}
 \begin{subfigure}{0.24\textwidth}
   \includegraphics[width=\linewidth]{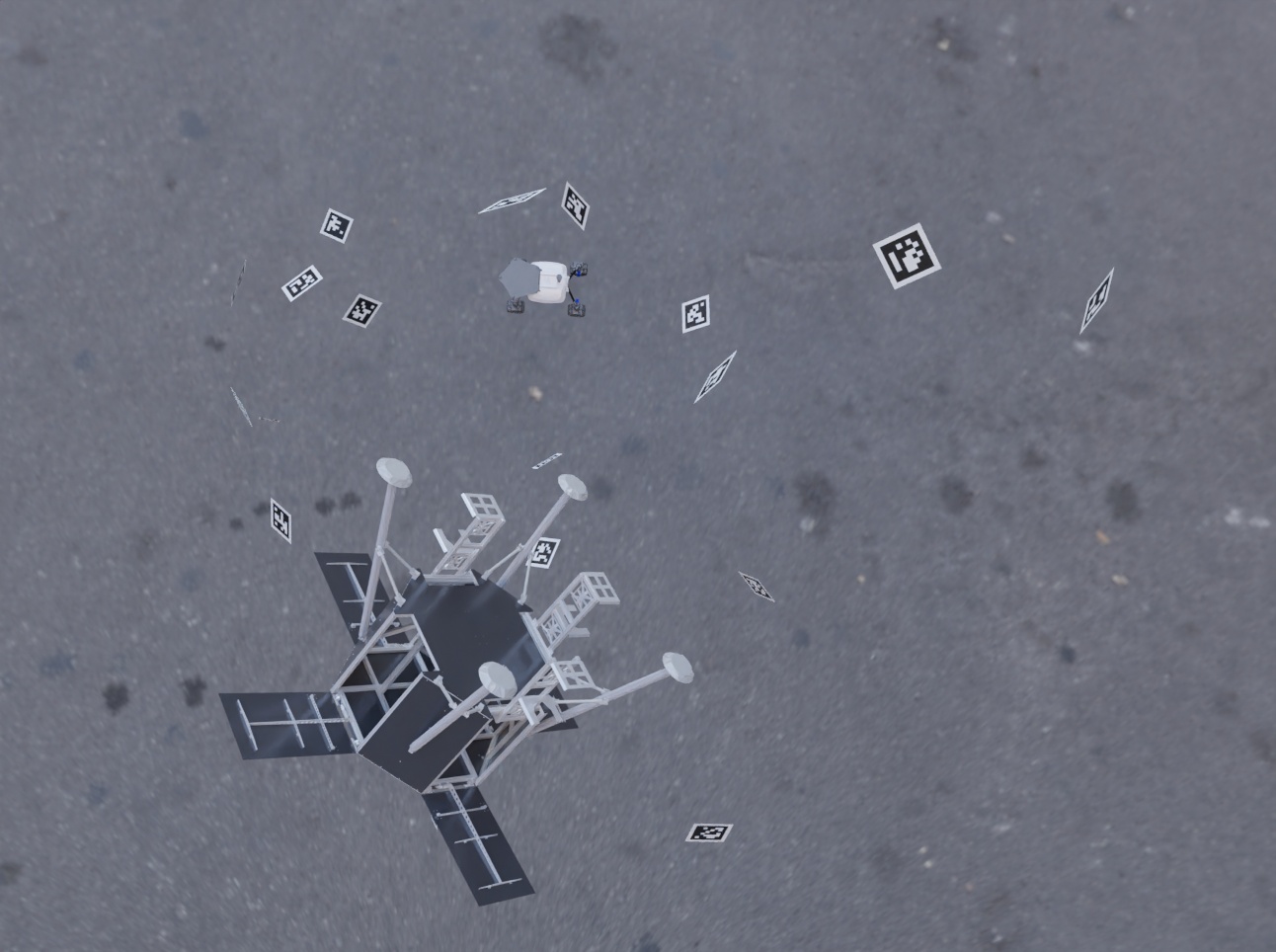}
 \end{subfigure}\\

 \vspace{0.4em}

    \begin{subfigure}{0.24\textwidth}
    \includegraphics[width=\linewidth]{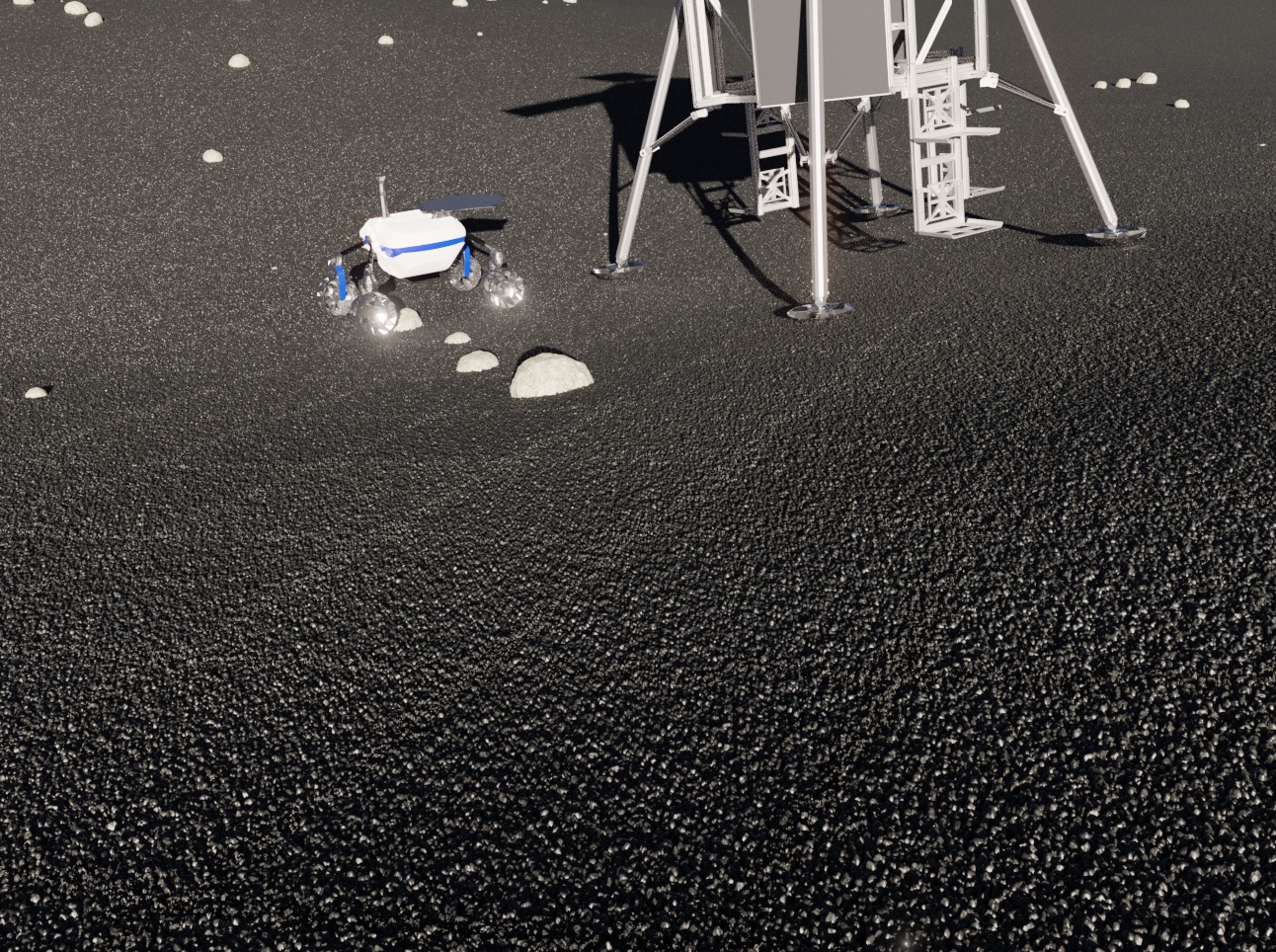}
  \end{subfigure}\hspace{0.1em}
  \begin{subfigure}{0.24\textwidth}
    \includegraphics[width=\linewidth]{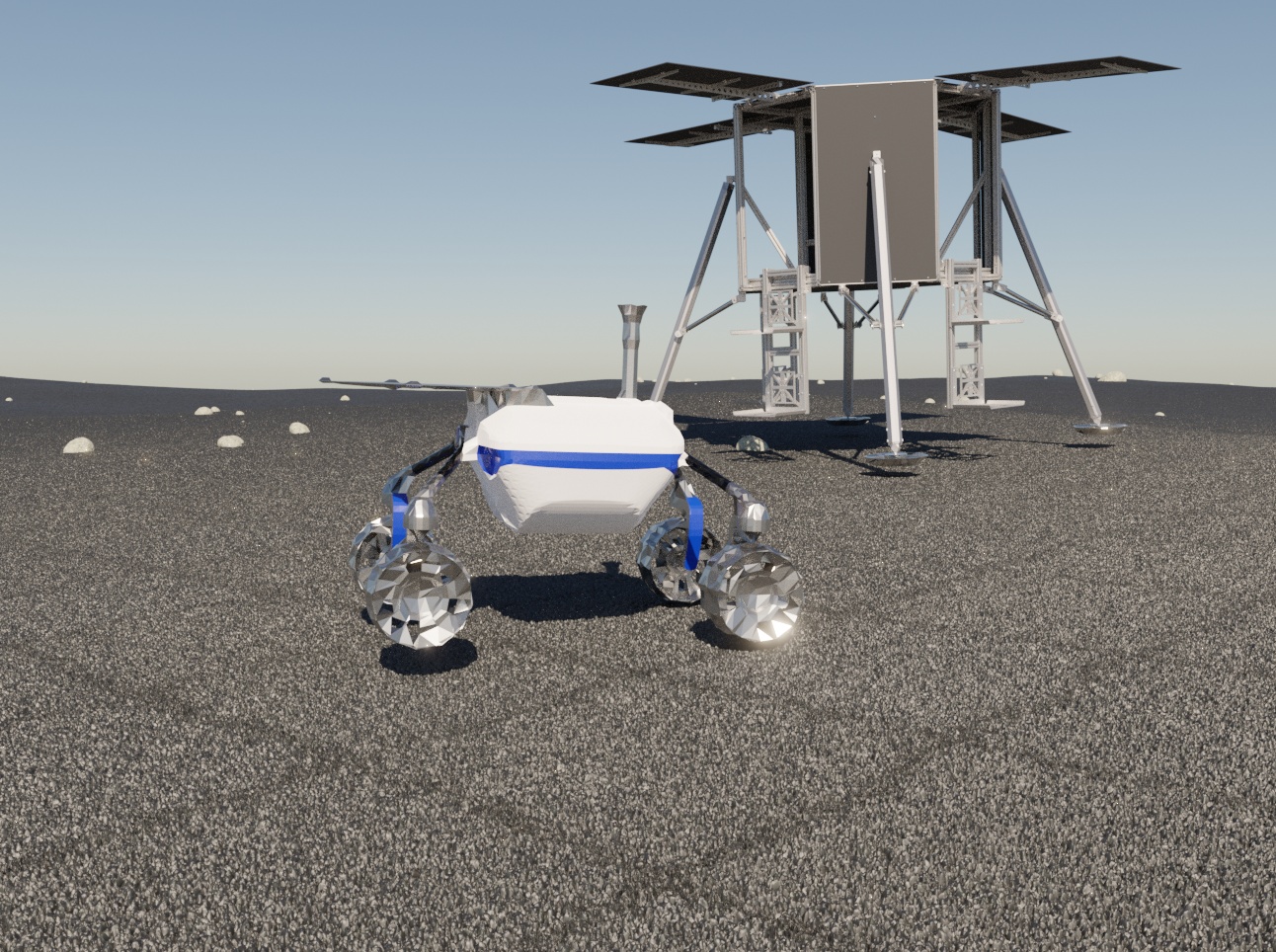}
  \end{subfigure}\hfill
 \begin{subfigure}{0.24\textwidth}
   \includegraphics[width=\linewidth]{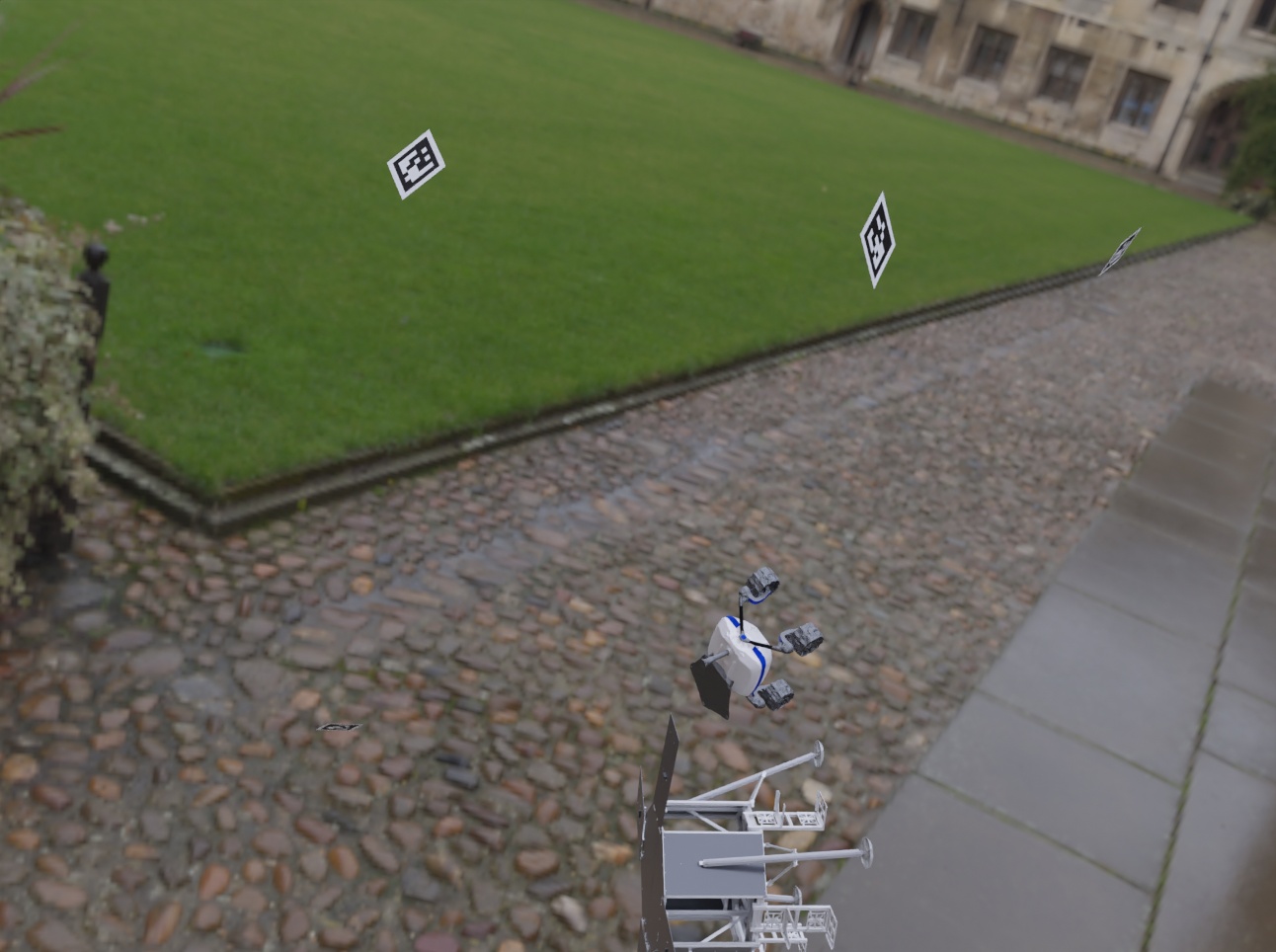}
 \end{subfigure}\hspace{0.1em}
 \begin{subfigure}{0.24\textwidth}
   \includegraphics[width=\linewidth]{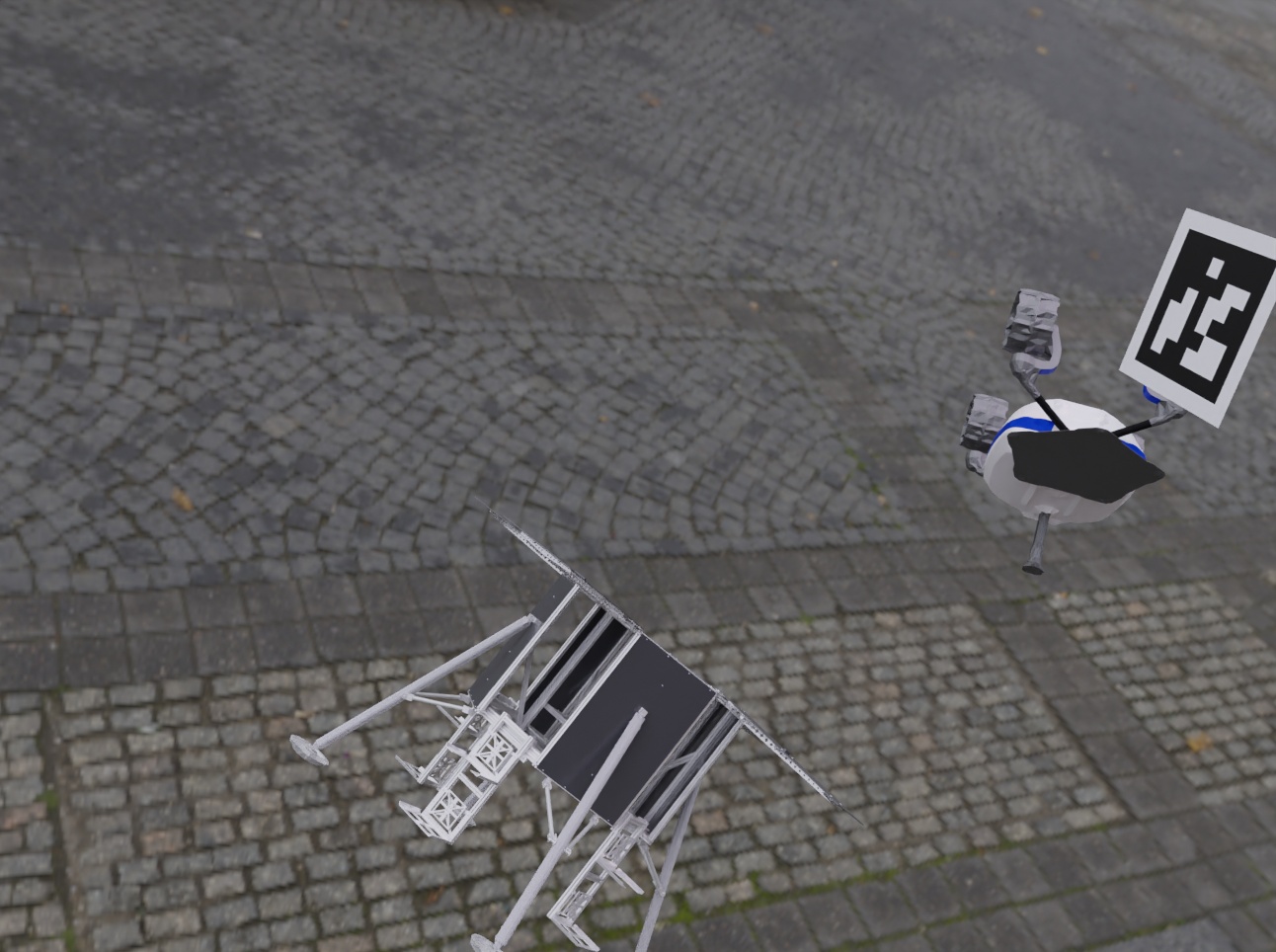}
 \end{subfigure}
\caption{Impressions of training samples. (\textbf{left:}) OAISYS images including LRU and Lander in an Etna-like setting. (\textbf{right:}) BlenderProc images, with LRU, Lander and random AprilTags for distractions, on a random background.}
  \label{fig:training_samples}
\end{figure*}

\subsection{Multi-Robot SLAM}\label{sec:multirobot_slam}
This work is based upon a decentralized multi-robot SLAM system, developed at the 
[Anonymous Institution]
for a team of heterogeneous robotic systems, such as the Lightweight Rover Unit (LRU) \cite{schuster2019towards} and the ARDEA UAV \cite{lutz2020ardea}, and described in detail in \cite{schuster2019distributed}. 

The SLAM system, in principle agnostic to the perception modality, is specialized for robots equipped with stereo vision. Local state estimation is computed as a Local Reference Filter \cite{schmid2014local} and fuses Visual Odometry (VO), IMU measurements, and odometric sources. State estimation is therefore referenced with respect to local frames that switch to the current estimated position when the pose covariance grows over a certain limit, marginalizing past states. Frame switches signal the creation of new submaps, which partition the environment in rigid elements. A submap matching modules attempts to register overlapping submaps to create loop closures constraints \cite{brand2015submap}. Similarly, keyframes are extracted from camera views, and AKAZE features are used to insert visual words in a growing database and recalling matching keyframes, in order to establish visual loop closures. 
\begin{figure}[!tp]
    \centering
    \includegraphics[width=\linewidth]{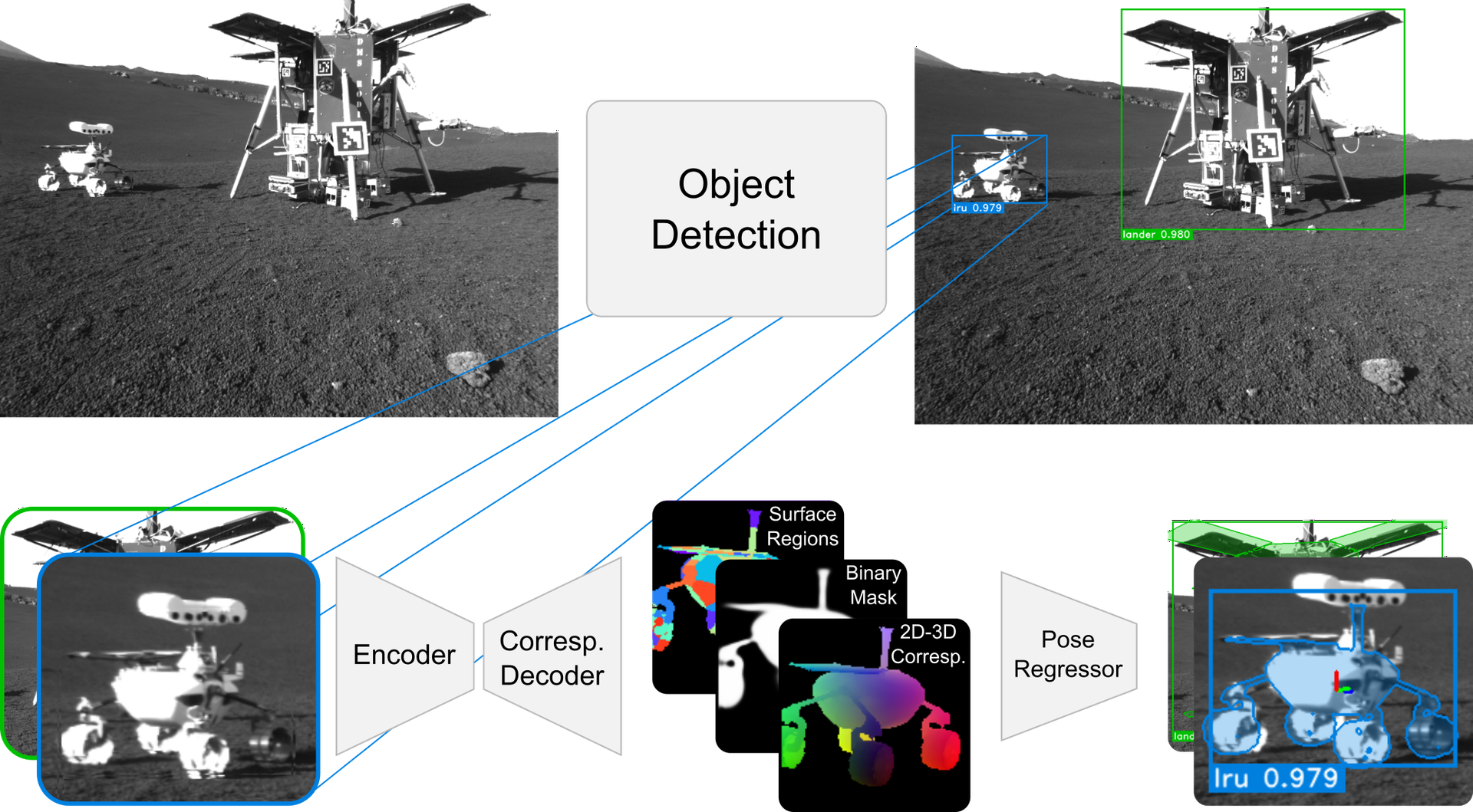}
    \caption{
    Illustration of the markerless detection pipeline. A 2D object detection first generates an image crop, which is then processed by the pose estimator. The encoder-decoder architecture produces three outputs: 2D–3D correspondences, foreground-background masks, and surface regions as an auxiliary task. The 2D–3D correspondences are subsequently passed to a pose regression network to predict the object’s 6D pose.
    }
    \label{fig:lru_pose_estimation}
\end{figure}

In order to join estimates from multiple robots into a combined SLAM session, correspondences between robot states are estimated either from visual loop closures or direct inter-robot pose measurements. The detection of an array of fiducial markers (AprilTag), mounted on the robots, provides a 6D transformation between the robot's bodies, that initially aligns the individual map frames and subsequently provides additional robust loop closure constraints. 
Frame switch transformations, loop closure and robot detection constraints, from all the robots operating during a mapping session, are inserted in a pose graph in a decentralized manner. The iSAM2 algorithm, from the GTSAM library, computes then updated poses at quasi-constant computational effort. An architectural overview of the multi-robot SLAM system is shown in Fig.~\ref{fig:architecture}.

\subsection{Markerless Robot Detection Pipeline}\label{sec:pose_est}
\begin{figure}
    \centering
    \includegraphics[width=\linewidth, trim=0.02cm 0 0.02cm 0.02cm, clip]{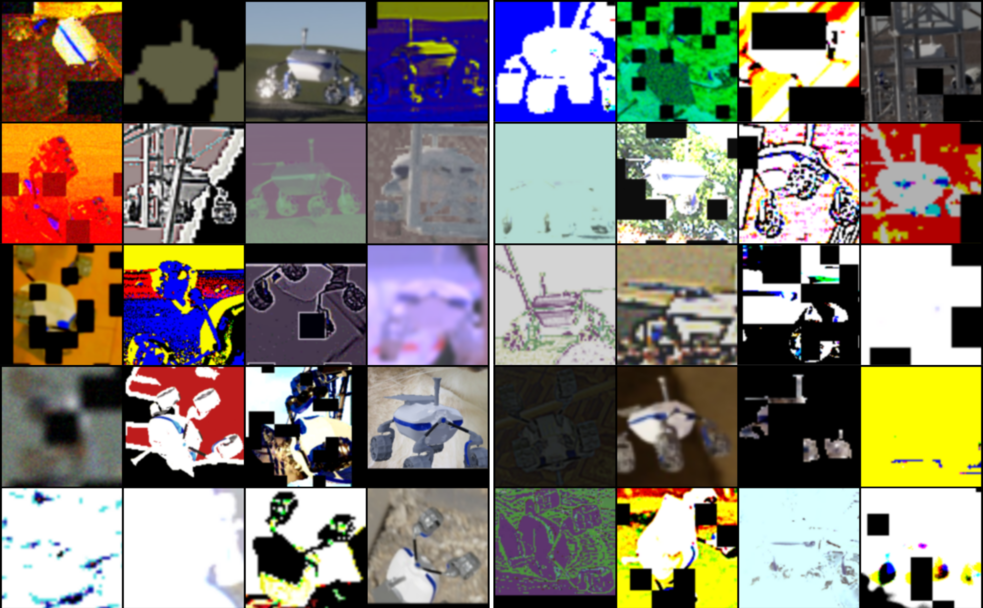}
    \caption{Examples of heavy augmentations on LRU instances from the synthetic samples used for an effective training of the pose estimation network in Sec.~\ref{sec:pose_est}}
    \label{fig:augmented_samples}
\end{figure}
In this section we present our proposed markerless robot detection and pose estimation component, depicted in Fig.~\ref{fig:lru_pose_estimation}.
\subsubsection{Detection and Instance Filtering} \label{sec:robot_detection}
Known robot instances are searched for in all input images using the YOLO v7 object detector.  
Detection results are ordered by confidence, and only the one with the highest confidence value is retained for further processing.
Excessive visual occlusion or cases where the robot are cropped by image borders are identified by checking how the aspect ratio of the bounding boxes differ to a value of 1. 
Only detections whose bounding boxes have aspect ratios within a specified interval \(\frac{1}{r} \leq \frac{height}{width} \leq r\) for a predefined aspect ratio \(r > 1\) are accepted.
Additionally, bounding boxes are discarded if they touch more than one image border, as in this case the aspect ratio check is insufficient to exclude partial detections.

\subsubsection{Pose Estimation}
We build upon the extended approach of Ulmer~\etal~\cite{ulmer20236d}.
It has demonstrated promising results in the domain of satellite pose estimation, a field that, like planetary exploration, has to deal with extreme lighting conditions.
Our experiments showed that the method can be efficiently trained on modern hardware in an end-to-end fashion.
Furthermore, it has sufficient capacity to handle the challenging conditions typical to space environments~\cite{ulmer_2025_aero} and also shows promising performance on other benchmark datasets, confirming its suitability for our application.
The approach employs a dense 2D-to-3D correspondence predictor, allowing the regression of 3D model coordinates for every pixel in the input image.
To further improve robustness, we introduce several adaptations.
We replace the decoder with a transformer-based architecture incorporating an attention mechanism.
Inspired by GDR-Net~\cite{Wang_2021_CVPR}, we substitute the classical PnP procedure with a neural network that directly regresses the target object's 6D pose.
Additionally, in parallel with the standard modal instance mask prediction, we introduce a new prediction head to estimate amodal masks, enhancing the system’s understanding of occluded object regions.

\begingroup

\definecolor{ValuePurple}{rgb}{0.2, 0.1, 1.0}
\definecolor{ValueBlue}{rgb}{0.2, 0.6, 0.8}
\definecolor{ValueGreen}{rgb}{0.2, 0.8, 0.3}

\colorlet{ValueGreen}{ValueBlue}
\colorlet{ValuePurple}{ValueBlue}


\colorlet{LightPurple}{ValuePurple!11}  \colorlet{LightBlue}{ValueBlue!13}  \colorlet{LightGreen}{ValueGreen!13}
\colorlet{MidPurple}{ValuePurple!32}    \colorlet{MidBlue}{ValueBlue!32}    \colorlet{MidGreen}{ValueGreen!33}
\colorlet{DarkPurple}{ValuePurple!52}   \colorlet{DarkBlue}{ValueBlue!58}   \colorlet{DarkGreen}{ValueGreen!59}   

\setlength{\tabcolsep}{1.5pt}
\renewcommand{\arraystretch}{1.9}
\setlength{\extrarowheight}{2.5pt}

\begin{table}[tp!]
    \centering
    \captionsetup{justification=centering}
    \caption{Synthetic Test Results: Detection rate $\rho$, translation error $t$, rotational error $\theta$ for different training and test set combinations.
    The arrow after a variable indicates whether a higher ($\uparrow$) or lower ($\downarrow$) value is desirable.}
    \label{tab:results:simulation_results_order1_columnformat}
    \renewcommand{\arraystretch}{1}
    \resizebox{\linewidth}{!}{
    \begin{tabularx}{\linewidth}{|c|c|c|Y|Y|Y|}
        \hline
        \footnotesize
        \multirow{2}{*}{\makecell{\scriptsize{\textbf{Test}} \\ \scriptsize{\textbf{Set}}}} & \multirow{2}{*}{\makecell{\scriptsize{\textbf{Performance}} \\ \scriptsize{\textbf{Metric}}}} & \multirow{2}{*}{\makecell{\scriptsize{\textbf{Observed}} \\ \scriptsize{\textbf{Robot}}}} & \multicolumn{3}{c|}{\textbf{Training Set}} \\
        \hhline{~~~---}
        & & & \scriptsize{\textbf{BPROC}} & \scriptsize{\textbf{OAISYS}} & \scriptsize{\textbf{COMBINED}} \\
        \hline
        \multirow{6}{*}{BPROC} & \multirow{2}{*}{$\rho\ \ \ \ \ \ \ \ (\uparrow)$} & Lander & \cellcolor{LightPurple}0.160 & \cellcolor{MidPurple}0.260 & \cellcolor{DarkPurple}\textbf{0.284} \\ \hhline{~~----}
        & & LRU & \cellcolor{MidPurple}0.080 & \cellcolor{LightPurple}0.064 & \cellcolor{DarkPurple}\textbf{0.324} \\ \hhline{~-----}
        & \multirow{2}{*}{$t\ \ \ [m]\ \ (\downarrow)$} & Lander & \cellcolor{MidBlue}0.310 & \cellcolor{LightBlue}1.497 & \cellcolor{DarkBlue}\textbf{0.259} \\ \hhline{~~----}
        & & LRU & \cellcolor{DarkBlue}\textbf{0.082} & \cellcolor{LightBlue}1.621 & \cellcolor{MidBlue}0.153 \\ \hhline{~-----}
        & \multirow{2}{*}{$\theta\ [rad]\ (\downarrow)$} & Lander & \cellcolor{DarkGreen}\textbf{0.421} & \cellcolor{LightGreen}2.207 & \cellcolor{MidGreen}0.514 \\ \hhline{~~----}
        & & LRU & \cellcolor{DarkGreen}\textbf{0.099} & \cellcolor{LightGreen}1.143 & \cellcolor{MidGreen}0.188 \\ \hline
        \multirow{6}{*}{OAISYS} & \multirow{2}{*}{$\rho\ \ \ \ \ \ \ \ (\uparrow)$} & Lander & \cellcolor{LightPurple}0.080 & \cellcolor{MidPurple}0.676 & \cellcolor{DarkPurple}\textbf{0.792} \\ \hhline{~~----}
        & & LRU & \cellcolor{LightPurple}0.020 & \cellcolor{MidPurple}0.184 & \cellcolor{DarkPurple}\textbf{0.764} \\ \hhline{~-----}
        & \multirow{2}{*}{$t\ \ \ [m]\ \ (\downarrow)$} & Lander & \cellcolor{LightBlue}0.184 & \cellcolor{MidBlue}0.044 & \cellcolor{DarkBlue}\textbf{0.040} \\ \hhline{~~----}
        & & LRU & \cellcolor{DarkBlue}\textbf{0.034} & \cellcolor{LightBlue}0.115 & \cellcolor{MidBlue}0.098 \\ \hhline{~-----}
        & \multirow{2}{*}{$\theta\ [rad]\ (\downarrow)$} & Lander & \cellcolor{LightGreen}0.483 & \cellcolor{MidGreen}0.089 & \cellcolor{DarkGreen}\textbf{0.048} \\ \hhline{~~----}
        & & LRU & \cellcolor{LightGreen}0.128 & \cellcolor{MidGreen}0.092 & \cellcolor{DarkGreen}\textbf{0.052} \\ \hline
    \end{tabularx}}
\end{table}
\endgroup

\begin{figure}[tp]
    \centering
    \begin{subfigure}[t]{0.49\columnwidth}
        \centering
        \includegraphics[width=\textwidth, trim={1.5cm 2cm 1.5cm 1.5cm}, clip]{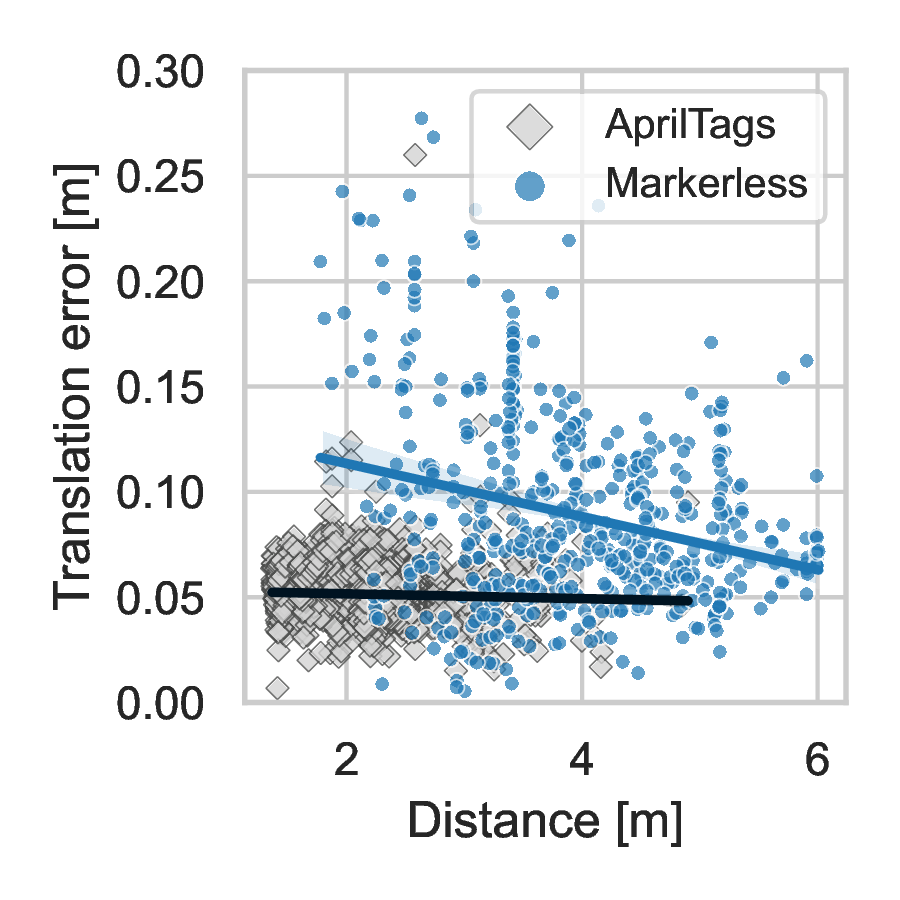}
        \phantomcaption
        \label{fig:first}
    \end{subfigure}
    \hfill
    \begin{subfigure}[t]{0.49\columnwidth}
        \centering
        \includegraphics[width=\textwidth, trim={1.5cm 2cm 1.5cm 1.5cm}, clip]{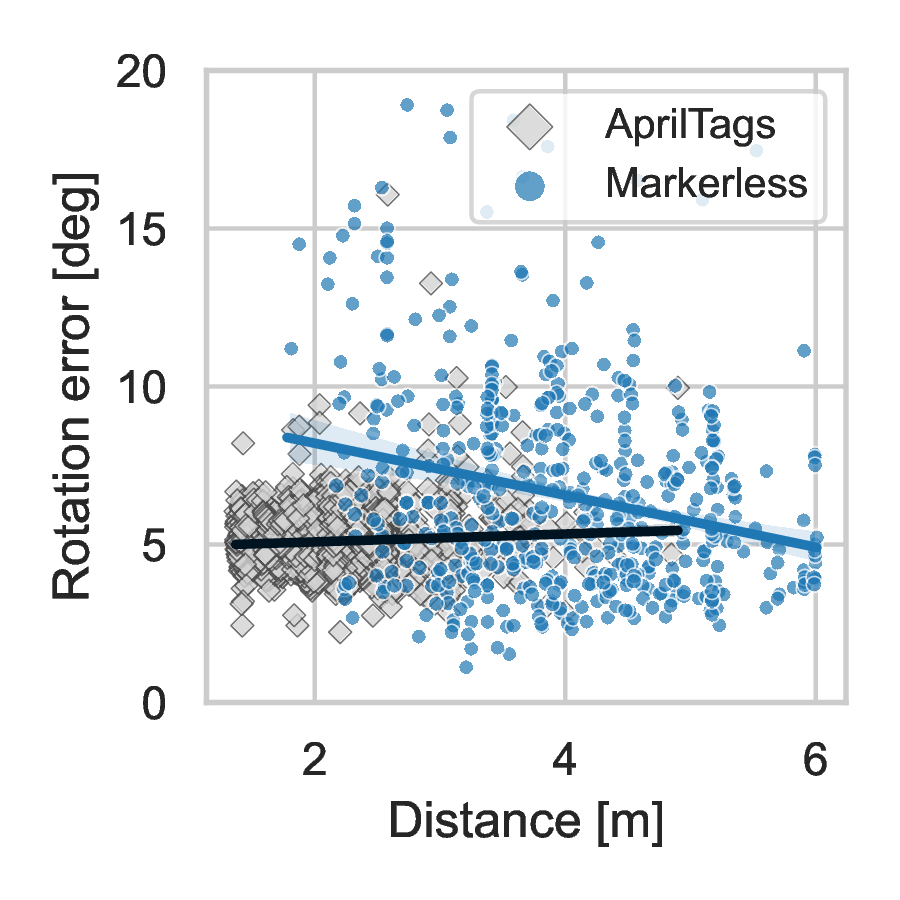}
        \phantomcaption
        \label{fig:second}
    \end{subfigure}
    \\
    \begin{subfigure}[t]{0.49\columnwidth}
        \centering
        \includegraphics[width=\textwidth]{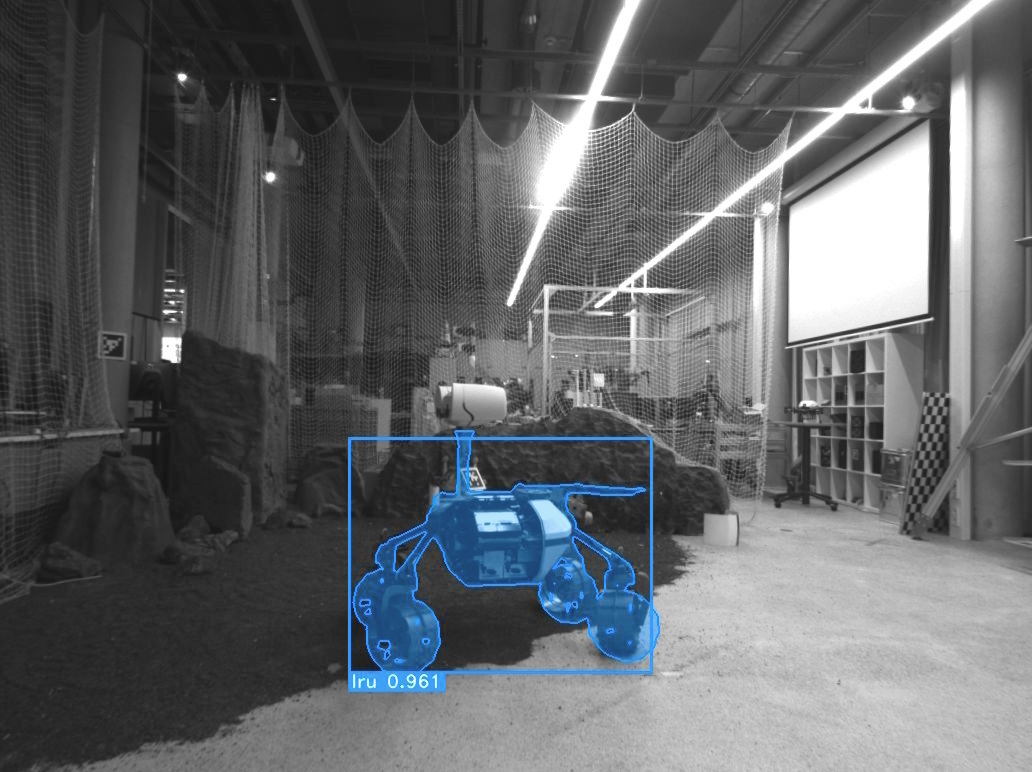}
        \phantomcaption
        \label{fig:ex1}
    \end{subfigure}
    \hfill
    \begin{subfigure}[t]{0.49\columnwidth}
        \centering
        \includegraphics[width=\textwidth]{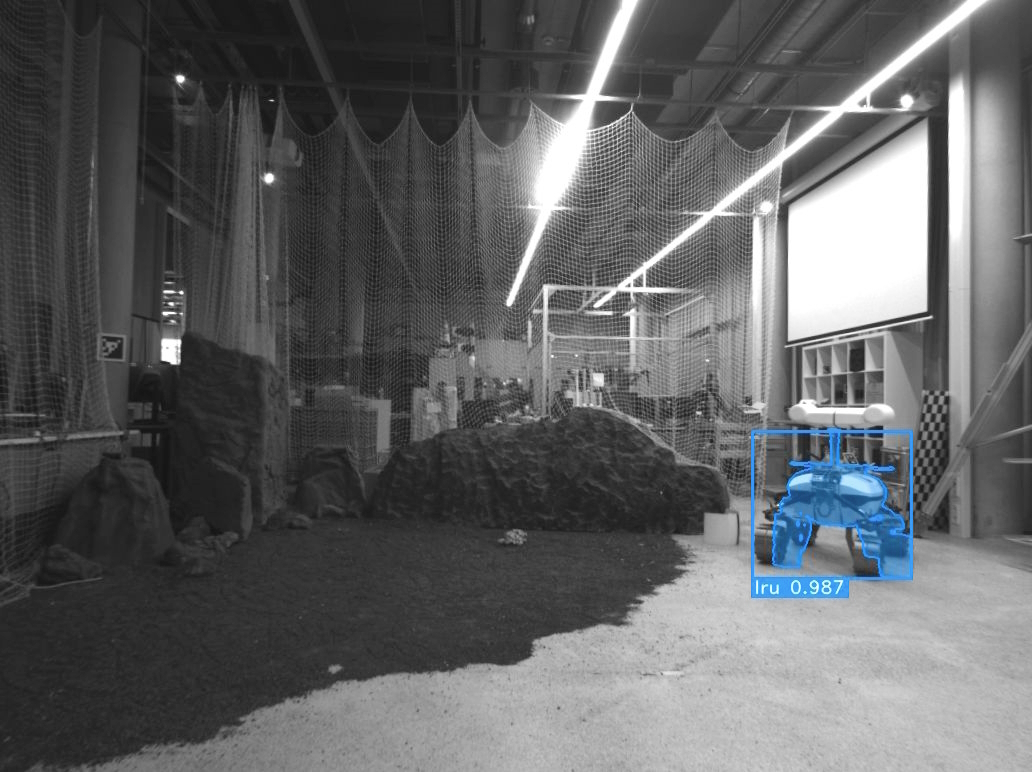}
        \phantomcaption
        \label{fig:ex2}
    \end{subfigure}
    \caption{Evaluation of translation (\textbf{top-left}) and rotation (\textbf{top-right}) errors of the proposed markerless pose estimation approach, and comparison with AprilTag detection, between the real LRU system, with LRU2 as observer, under VICON ground-truth measurements. (\textbf{bottom:}) Visual examples of markerless robot detection and re-projection of the CAD model on the image.}
    \label{fig:vicon_eval}
\end{figure}

\subsection{Training on Synthetic Data} \label{sec:training_process}

Both models for instance detection (Sec.\ref{sec:robot_detection}) and pose estimation (Sec.\ref{sec:pose_est}) were trained on synthetic data generated from simplified CAD models of the known robot types.
Both variants of the Lightweight Rover Unit, LRU and LRU2, were treated as a single type of robot, with training performed on an intersection of their respective 3D geometries.
A dataset was generated containing \(8000\) photo-realistic image samples, each annotated with binary object masks and reporting ground truth 6D poses with respect to the virtual camera.
The first half of the training samples were generated using DLR's open-source rendering framework BlenderProc~2~\cite{Denninger2023}, a Blender-based procedural pipeline to generate synthetic datasets, providing photo-realistic images and ground-truth annotations such as depth and object masks. To enhance robustness, BlenderProc supports generation of distractor objects and random 360°-backgrounds from both indoor and outdoor environments.
The second half of the training samples were generated using the open-source rendering and simulation framework OAISYS~\cite{Mueller2021}, a Blender-based procedural simulator for unstructured outdoor environments with a specific focus on simulating terrains and natural scenes. Lastly, as previously experienced in \cite{ulmer_2025_aero}, we implement a variety of strong data augmentations during training on synthetic input images, see Fig.~\ref{fig:augmented_samples}, in order to enhance robustness to challenging visual conditions for real image samples. 

\section{EVALUATION AND DISCUSSIONS}
\subsection{Evaluation of Pose Estimation on Synthetic Data}\label{sec:train}
Markerless detection and pose estimation performances were first validated on synthetic data using a desktop computer equipped with an Intel i9 and an NVIDIA RTX4000.
Different models were trained separately on BlenderProc data, OAISYS data and combined data from both sources.
The separate models were then evaluated on BlenderProc and OAISYS test datasets.
The test set included 500 unseen BlenderProc and OAISYS samples, for which detection rates and pose accuracy were measured. 
Table~\ref{tab:results:simulation_results_order1_columnformat} reports detection rate, translation and rotation accuracy of each model on two separate BlenderProc- and OAISYS-only test data, showing that the best performances are achieved on training with diverse data. For a collection of visual impressions of generated data from both simulation toolkits, see Fig.~\ref{fig:training_samples}. 
In this evaluation, we measured the runtime of the combined detection and pose estimation steps achieving rates of up to 14 Hz, while for the detection alone up to 20 Hz.  

\begingroup
\setlength{\tabcolsep}{1.8pt}
\begin{table}[!tp]
\caption{Multi-robot detection metrics during the SLAM tests comparing the baseline approach with AprilTag only, and the proposed addition of Markerless detection and pose estimation}
\label{tab:combined_slam_error}
\centering
\renewcommand{\arraystretch}{1.2}
\resizebox{\linewidth}{!}{
\begin{tabular}{lccc}
\hline
\textbf{Performance Metric} & \textbf{Tag} & \makecell{\textbf{Tag +} \\ \textbf{Markerl.}} & \textbf{Improvement}\\
\hline
\multicolumn{1}{l}{\textbf{Mission 1, LRU}} \\
\hhline{----}
\# Detections               & 49 & 87 & \dgreen{\textbf{+87\%}} \\ 
Max. Det. Distance [m]      & 4.94 & 16.15 & \dgreen{\textbf{+227\%}} \\
Max. Open-Loop Duration [s] & 2300 & 1033 & \dgreen{\textbf{-55\%}} \\
Traj. Error [m]             & 1.57 & 1.13 & \dgreen{\textbf{-28\%}} \\ 
\hline
\multicolumn{1}{l}{\textbf{Mission 1, LRU2}} \\
\hhline{----}
\# Detections               & 154 & 180 & \dgreen{\textbf{+17\%}} \\ 
Max Det. Distance [m]       & 5.63 & 9.12 & \dgreen{\textbf{+62\%}} \\
Max Open-Loop Duration [s]  & 980 & 608 & \dgreen{\textbf{-38\%}}\\
Traj. Error [m]             & 0.75 & 0.78 & \dred{\textbf{+4\%}} \\ 
\hline
\multicolumn{1}{l}{\textbf{Mission 2, LRU}} \\
\hhline{----}
\# Detections               & 14 & 20 & \dgreen{\textbf{+43\%}}\\ 
Max. Det. Distance [m]      & 3.42 & 17.14 & \dgreen{\textbf{+402\%}}\\
Max. Open-Loop Duration [s] & 440 & 203 & \dgreen{\textbf{-54\%}}\\
Traj. Error [m]             & 0.76 & 0.52 & \dgreen{\textbf{-32\%}}\\ 
\hline
\multicolumn{1}{l}{\textbf{Mission 2, LRU2}} \\
\hhline{----}
\# Detections               & 18 & 19 & \dgreen{\textbf{+6\%}}\\
Max. Det. Distance [m]      & 7.63 & 7.63 & \textbf{--}\\ 
Max. Open-Loop Duration [s] & 490 & 490 & \textbf{--}\\
Traj. Error [m]             & 0.96 & 1.05 & \dred{\textbf{+9\%}}\\
\hline
\end{tabular}}
\end{table}
\endgroup

\subsection{Evaluation of Pose Estimation on Real Data}
\begin{figure*}[tp!]
    \centering
    \resizebox{\textwidth}{!}{
    \begin{tabular}{l r}
    
    \hspace{-1.5ex}\multirow{2}{.51\textwidth}{
    \begin{subfigure}[t]{0.521\textwidth}
    \centering
        \includegraphics[width=\textwidth]{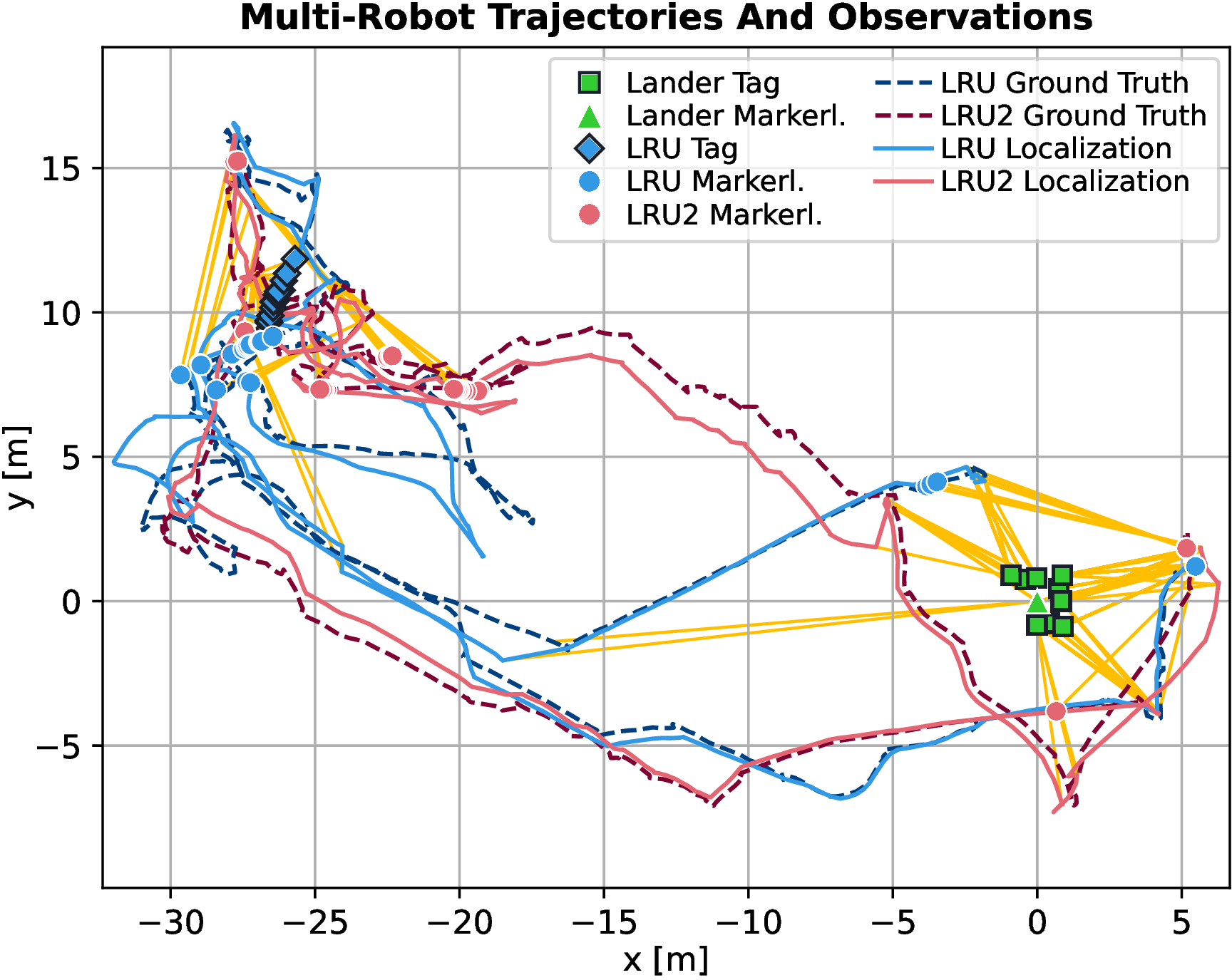}
        \phantomcaption
        \label{fig:mission1_trajectory}
    \end{subfigure}\hfill}\\[0.33ex]
    & \hfill \begin{subfigure}[t]{0.444\textwidth}
        \centering
        \vspace{-3.25mm}
        \includegraphics[width=\textwidth]{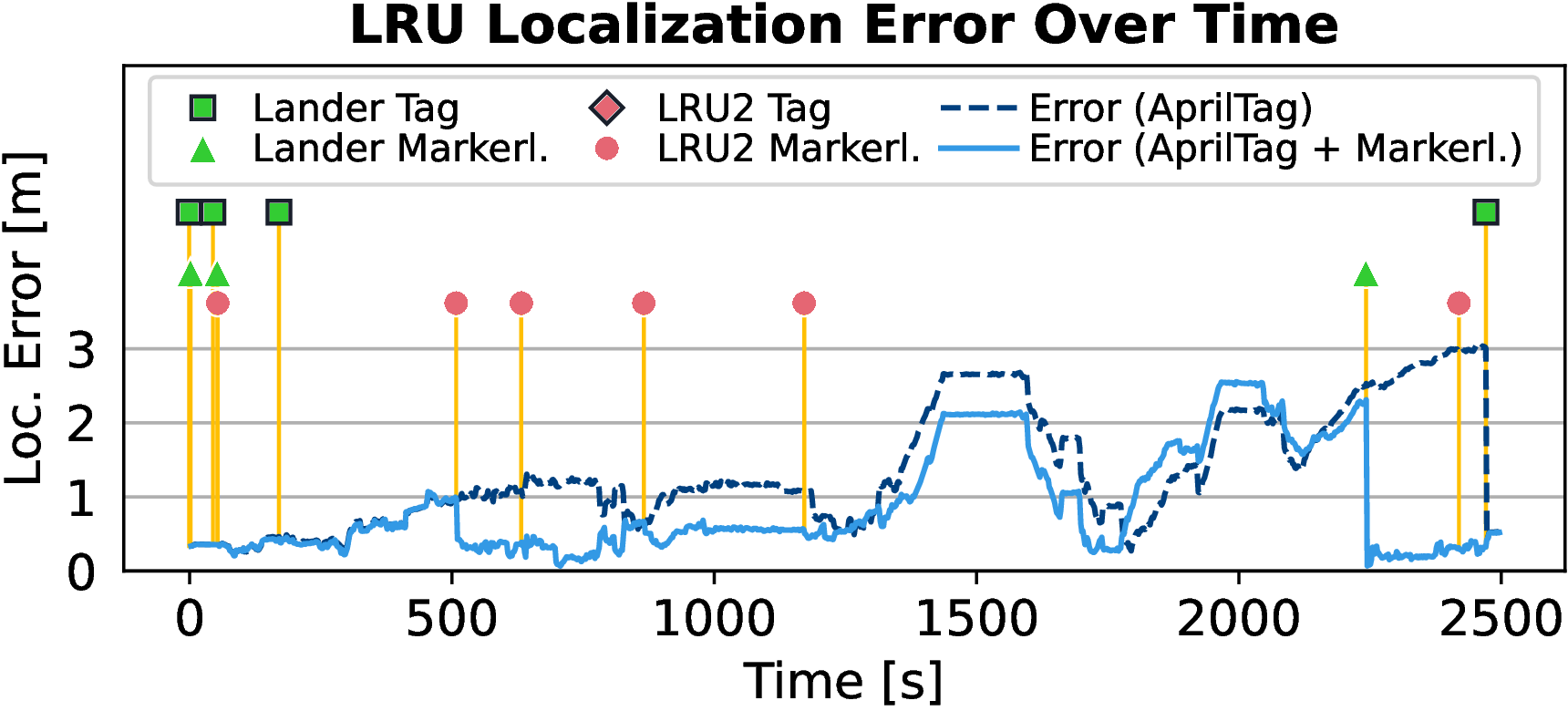}
        \phantomcaption
        \label{fig:mission1_lru}
    \end{subfigure} \\ [-1.5ex]
    &\hfill \begin{subfigure}[t]{0.444\textwidth}
        \centering
        \includegraphics[width=\textwidth]{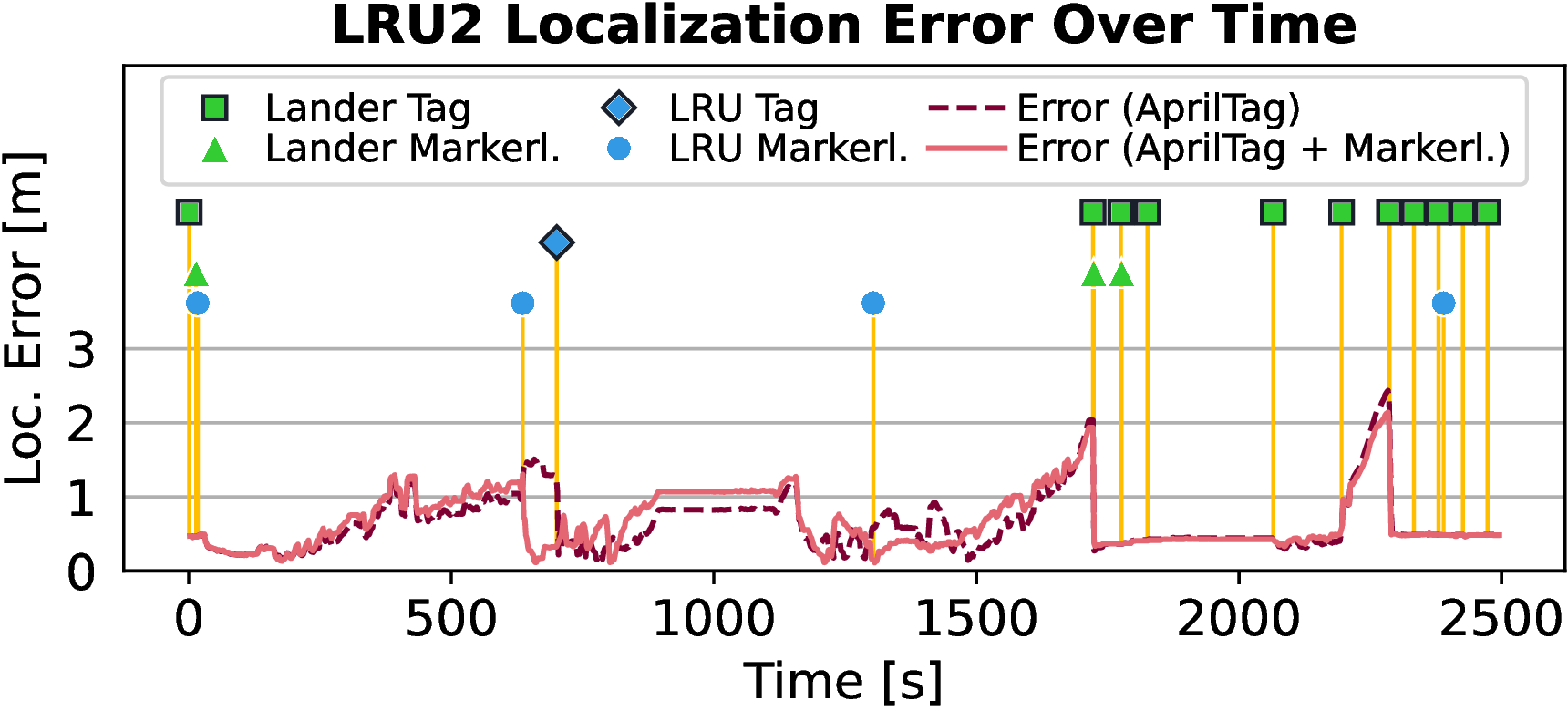}
        \phantomcaption
        \label{fig:mission1_lru2}
        \end{subfigure} 
    \end{tabular}}

    \caption{Multi-robot SLAM results from \textbf{Mission 1}. (\textbf{left:})  Trajectory of the two rovers over the entire mission, localized via visual multi-robot SLAM using full input knowledge. (\textbf{top right:}) Localization error of LRU, when observing the Lander and LRU2. Error correction after markerless LRU2 observations at \textbf{516s}, \textbf{874s} and markerless Lander detection at \textbf{2250s}. (\textbf{bottom right:}) Localization error of LRU2, when observing the Lander and LRU. Error correction after markerless LRU detections at \textbf{636s}, \textbf{1304s}.}
    \label{fig:real_world_slam_eval_mission_1}
\end{figure*}
\begin{figure*}[!ht]
    \centering
    \resizebox{\textwidth}{!}{
    \begin{tabular}{l r}

    \hspace{-1.5ex}\multirow{2}{.51\textwidth}{
    \begin{subfigure}[t]{0.49\textwidth}
    \centering
        \includegraphics[width=\textwidth]{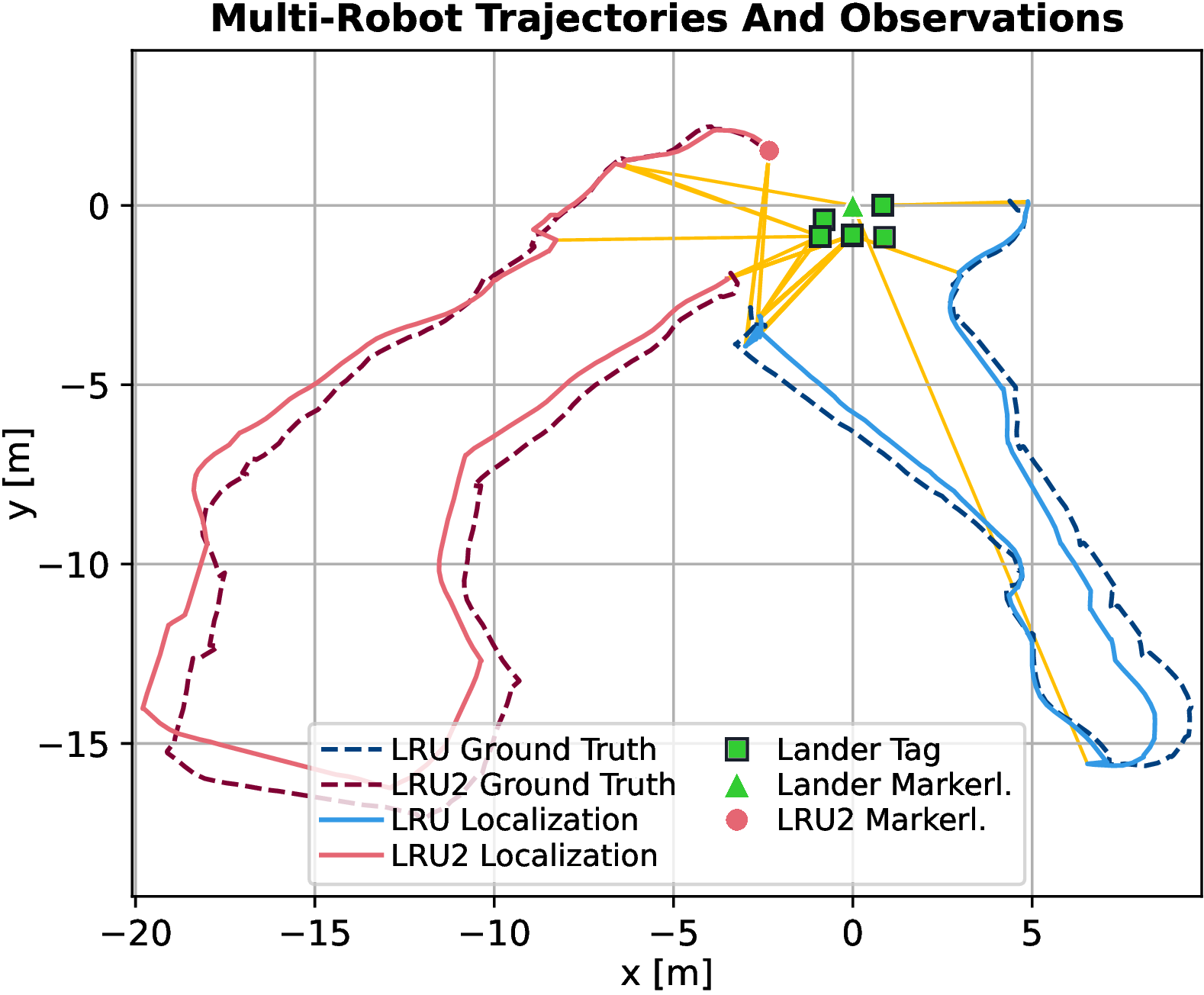}
        \phantomcaption
        \label{fig:mission2_trajectory}
    \end{subfigure}\hfill}
    & \hfill \begin{subfigure}[t]{0.444\textwidth}
        \centering
        \vspace{-2.5mm}
        \includegraphics[width=\textwidth]{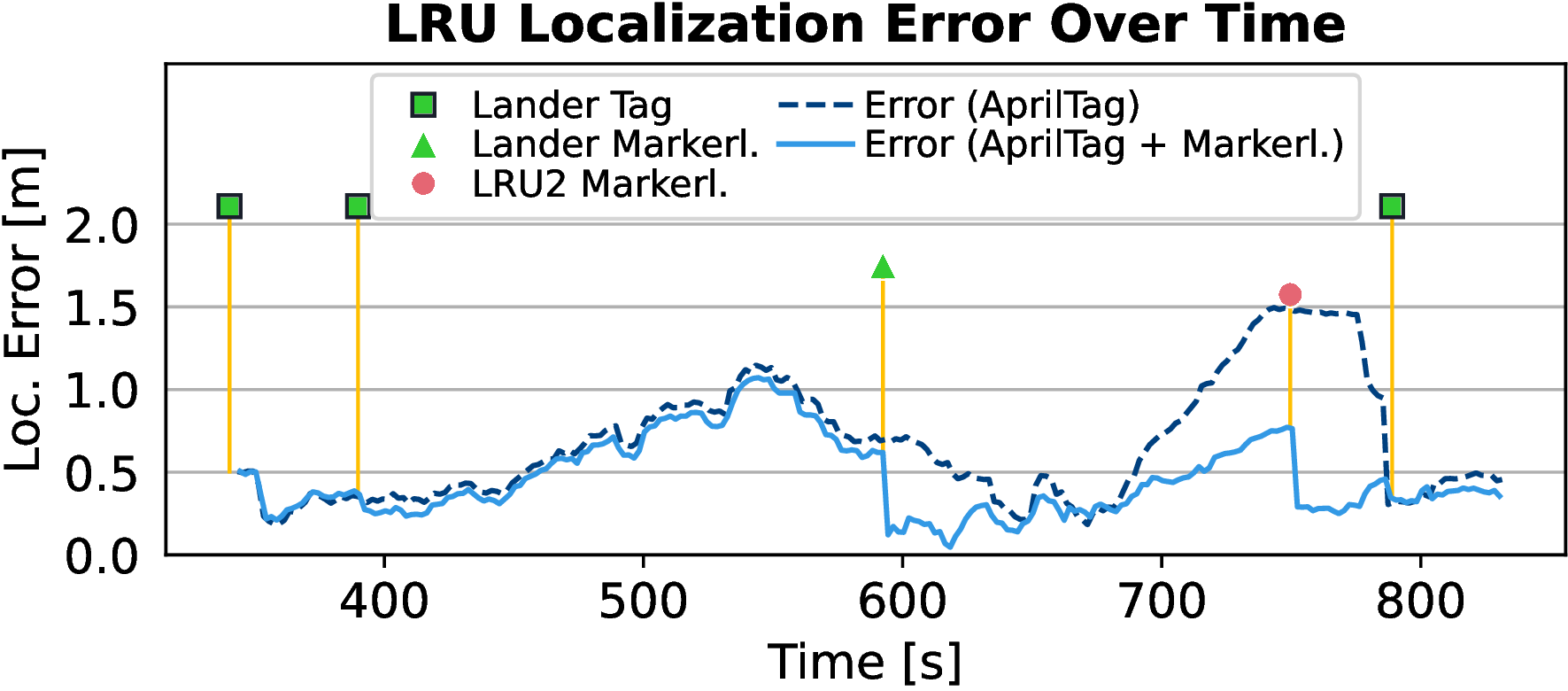}
        \phantomcaption
        \label{fig:mission2_lru}
    \end{subfigure} \\ [-1.5ex]
    &\hfill \begin{subfigure}[t]{0.444\textwidth}
        \centering
        \includegraphics[width=\textwidth]{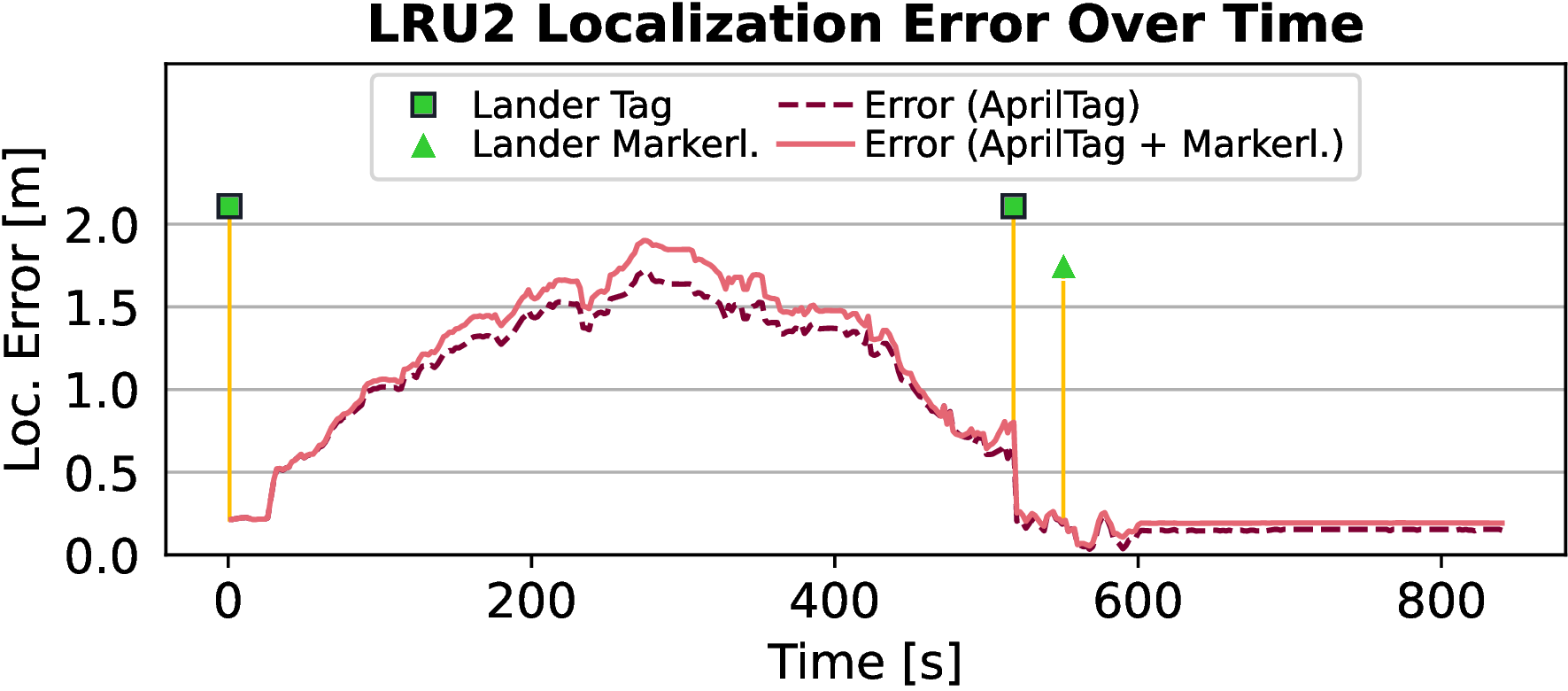}
        \phantomcaption
        \label{fig:mission2_lru2}
        \end{subfigure} 
    \end{tabular}}

    \caption{Multi-robot SLAM results from \textbf{ Mission 2}. (\textbf{left:})  Trajectory of the two rovers over the entire mission, localized via visual multi-robot SLAM using full input knowledge. (\textbf{top right:}) Localization error of LRU, when observing the Lander and LRU2. Error correction after markerless Lander observation at \textbf{593s} and markerless LRU2 observation at \textbf{751s}. (\textbf{bottom right:}) Localization error of LRU2, when observing the Lander and LRU. No significant differences with or without the markerless modality.}
    \label{fig:real_world_slam_eval_mission_2}
\end{figure*}
We now evaluate, on real image samples, the performances of the pose estimation approach trained in simulation (Sec.~\ref{sec:train}) using the generic LRU model without pan-tilt head. Both LRU systems, with LRU2 being the observer and LRU being the target, are used in an indoor laboratory settings equipped with a VICON system. With the observer placed in one corner of the VICON covered area, the target robot is manually driven in order to create a dataset of relative pose samples that cover as densely as possible ranges and relative orientations from the available space. The pose estimation method is then compared against a baseline multi-robot detection approach based on the detection of an array of AprilTags mounted around the neck of both rovers. Both the markerless- and AprilTag-based approaches estimate the transformation from the rover's body centers. Measurements are then compared with a VICON ground truth. Results in Fig.~\ref{fig:vicon_eval} highlight the distribution of rotation and translation errors separately against the true distance from the body centers. The first important insight is that the baseline approach fails to cover the entire range of distances, showing a distinct limit around the 4 meters range, which can vary with the marker size. Secondly, both translation and rotation accuracy of markerless pose estimations improves at larger distances. We assume that this behavior depends on a decreasing effect of perspective distortion at longer ranges, and may as well be compensate with carefully tuned synthetic data generation for training. Furthermore, the results highlight how both modalities complement each other with a clear benefit for inclusion in a SLAM system. 

\subsection{Evaluation in a Multi-Robot SLAM System}
To validate the integrated system performance of the markerless SLAM approach in a real-world use case, the proposed markerless robot detection approach was interfaced as a ROS2 component in the multi-robot SLAM solution described in Sec.~\ref{sec:multirobot_slam}.
The SLAM system was evaluated offline on a variety of data recorded during the 2022 ARCHES field test campaign \cite{schuster2020arches} on the analogous planetary surface of Mt. Etna, Sicily.
In this evaluation, we consider the recordings of two multi-robot navigation experiments, denoted here Mission 1 and Mission 2, that lasted respectively 42 min and 14 min, during which the LRUs drove a wide trajectory in a \(40\)m\(\times\)\(40\)m area around the Lander. The Lander, equipped with large AprilTags, functioned as the central reference for the multi-robot SLAM results. At the beginning of each mapping session, the LRUs glanced at the Lander to establish a connection to the elected global frame before proceeding to navigate to user-defined waypoints. 
Thanks to the distributed nature of the employed SLAM system, the LRUs communicated at all time any measurement that affected the structure of their shared SLAM graph, for instance AprilTag-based and markerless robot detections, visual loop closures, or additions of new submaps. For a detailed explanation of the SLAM system and multi-robot communication approach we refer to \cite{schuster2019distributed}.


This evaluation focuses on \textit{instantaneous} localization accuracy rather than traditional \textit{a-posteriori} metrics, as the former is what directly impacts the performance of open-loop navigation toward POIs. We measure multi-robot detection frequency and maximum range while quantifying the reduction of the length of open-loop navigation sequences, that is, navigation sequences between detection events. Localization performance is summarized as the RMSE (Root Mean Square Error) between SLAM estimates and D-GNSS measurements.

As shown in Table~\ref{tab:combined_slam_error}, our markerless approach significantly improves detection rates, ranges, and accuracy across both missions 
at the cost of potential minor fluctuations (see LRU2 trajectory error in Table~\ref{tab:combined_slam_error}) due to sub-par rotational accuracy of markerless multi-robot detection at very close ranges compared to fiducial markers, effect that is already reported in Fig.~\ref{fig:vicon_eval}. 
Finally, Fig.~\ref{fig:real_world_slam_eval_mission_1} and Fig.~\ref{fig:real_world_slam_eval_mission_2} visualize trajectories and their accuracy against a D-GNSS ground truth over mission time, highlighting sharp error reductions triggered by markerless detection events.

\section{CONCLUSIONS}
Markerless robot detection delivers a significant performance improvement in the context of multi-robot SLAM, especially when operating in perceptually challenging environments, overcoming the limitations imposed by traditional approaches based on fiducial markers (e.g., AprilTag), which are limited in range and sensitive to lighting conditions. For robots whose geometry is known as full or partial CAD models the proposed approach increases the distance of robot detection up to 400\% during our field tests, regardless of lighting conditions, leading to significant improvement on the instantaneous localization accuracy, which has paramount importance during live operations. Future work will explore solutions for robots as articulated ensembles of rigid components (head, torso, wheels, etc.), as well as the deployment of the approach into embedded GPUs for online operation.






\bibliographystyle{IEEEtran}
\bibliography{refs}

\end{document}